\documentclass[11pt,a4paper]{article}
\usepackage{times,latexsym}
\usepackage{url}
\usepackage[T1]{fontenc}
\usepackage{pifont} 
\usepackage{xcolor} 
\usepackage{float}   
\usepackage{graphicx}
\usepackage{subcaption}
\usepackage{colortbl}        
\usepackage{multirow}        
\usepackage{graphicx}
\usepackage{tcolorbox}
\usepackage[normalem]{ulem}
\usepackage{xurl}
\usepackage{hyperref}   
\usepackage{amsmath}
\usepackage{amsmath,amssymb}
\usepackage{amsfonts}

\usepackage[acceptedWithA]{tacl2021v1}
\usepackage[table]{xcolor}
\usepackage{pifont}
\usepackage{paralist}
\usepackage{booktabs}
\usepackage{siunitx}
\usepackage{xspace,mfirstuc,tabulary}

\tcbuselibrary{breakable}

\newif\iftaclinstructions
\taclinstructionsfalse
\iftaclinstructions
\renewcommand{\confidential}{}
\renewcommand{\anonsubtext}{(No author info supplied here, for consistency with
TACL-submission anonymization requirements)}
\newcommand{\instr}
\fi
\usepackage{makecell}
\usepackage{booktabs}
\usepackage{array}
\usepackage{colortbl}

\iftaclpubformat

\newcommand{\cmark}{\ding{51}}%
\newcommand{\xmark}{\ding{55}}%
\newcommand{\rpoint}{\ding{43}}%
\newcommand{\done}{\textcolor{green!70!black}{\cmark}}
\newcommand{\notdone}{\textcolor{red!70!black}{\xmark}}
\newcommand{\sysBL}[1]{\textsc{Bleu}}
\newcommand{\sysCHRF}[1]{\textsc{chrF}}
\newcommand{\sysMT}[1]{\textsc{Meteor}}
\newcommand{\sysRG}[1]{\textsc{Rouge}}
\newcommand{\sysSBERT}[1]{\textsc{S-Bert}}
\newcommand{\sysCSIM}[1]{\textsc{Cosine Similarity}}
\newcommand{\sysI}[1]{\textsc{IDEFICS}}
\newcommand{\sysIB}[1]{\textsc{InstructBLIP}}
\newcommand{\sysLV}[1]{\textsc{LLaVA}}
\newcommand{\sysGPT}[1]{\textsc{GPT}}
\newcommand{\sysGPTO}[1]{\textsc{GPT-4o}}
\newcommand{\sysLM}[1]{\textsc{LLaMA-Vision}}
\newcommand{\sysM}[1]{\textsc{Molmo}}
\newcommand{\sysP}[1]{\textsc{PaliGemma}}
\newcommand{\sysQ}[1]{\textsc{Qwen}}
\newcommand{\metchr}[1]{\textsc{chr-F1}}
\newcommand{\meteor}[1]{\textsc{Meteor}}
\newcommand{\metSS}[1]{\textsc{Semantic-Similarity}}
\newcommand{\fhm}[1]{\textbf{FHM}}
\newcommand{\mami}[1]{\textbf{MAMI}}

\newcommand{\ssubha}[1]{\textcolor{black}{#1}}
\newcommand{\cam}[1]{\textcolor{black}{#1}}

\newcommand{\ours}[1]{\textbf{\textsc{StemTox}}}
\newcommand{\datas}[1]{\textbf{\textsc{ToxicTags}}}
\fi

\title{\ours{}: 
From \cam{Collaborative} Tags to Fine-Grained Toxic Meme Detection via Entropy-Guided Multi-Task Learning}
\author{
{\bfseries
Subhankar Swain$^{*}$ \quad
Naquee Rizwan$^{*}$ \quad
Vishwa Gangadhar S} \\
{\bfseries
Nayandeep Deb \quad
Animesh Mukherjee} \\
Indian Institute of Technology, Kharagpur, India \\
\texttt{\{subhankar.swain25,nrizwan,vishwa2488,nayandeepdeb125\}@kgpian.iitkgp.ac.in} \\
\texttt{animeshm@cse.iitkgp.ac.in} \\
$^{*}$Equal contribution.
}

\date{}

\begin{document}
\maketitle
\begin{abstract}
Memes, as a widely used mode of online communication, often serve as vehicles for spreading harmful content. However, limitations in data accessibility and the high costs of dataset curation hinder the development of robust meme moderation systems. To address this challenge, in this work, we introduce a first-of-its-kind dataset -- \datas{} consisting of 6,300 real-world meme-based posts annotated in two stages: (i) binary classification into \textit{toxic} and \textit{normal}, and (ii) fine-grained labelling of toxic memes as \textit{hateful}, \textit{dangerous}, or \textit{offensive}. A key feature of this dataset is that it includes \cam{\textit{collaborative tags}} associated with the original posts, enhancing the context of each meme. In addition, we propose a novel entropy-guided multi-tasking framework -- \textbf{\ours{}} -- that leverages these \cam{collaborative tags} alongside visual and textual inputs within a robust classification framework. Experimental results show that incorporating these tags substantially enhances the performance of state-of-the-art VLMs in toxicity detection tasks. Our contributions offer a novel and scalable foundation for improved content moderation in multimodal online environments. We have made our code\footnote{Code: \url{https://github.com/hate-alert/STEMTOX}}
and dataset\footnote{Dataset: \url{https://huggingface.co/datasets/swainsubhankar/ToxicTags}}
publicly available for research purposes. \textit{\textbf{\textcolor{red}{Warning: Contains potentially toxic contents.}}}\end{abstract}

\section{Introduction}
\label{main:introduction}
While communication on online platforms span a variety of modalities\footnote{\url{https://www.sprinklr.com/blog/types-of-social-media/}}, memes have emerged as a popular and influential form of expression -- initially intended for lighthearted humor. However, they are increasingly being misused as vehicles for spreading harmful content, including hate speech, misinformation, and toxic ideologies.

\noindent\textbf{Model limitations:} Identifying such harmful content is particularly challenging due to the subtle and context-dependent nature of memes. Their meaning is often embedded in cultural references, online trends, sarcasm, or coded language, making them difficult to interpret not only for automated systems but even for human moderators. In response to these challenges, vision language models (VLMs) have recently gained traction as powerful tools for content moderation. These models are capable of jointly analysing visual and textual elements to grasp the nuanced context of memes and can provide detailed justifications for their classifications~\cite{qu2025meme}. Despite these advances, recent studies have highlighted the limitations of VLMs in accurately detecting hateful memes~\cite{rizwan2025exploring}, underscoring the urgent need to bridge these performance gaps. Similarly, a recent blog post\footnote{\url{https://about.fb.com/news/2025/01/meta-more-speech-fewer-mistakes/}} by Meta highlights the challenges and complexities inherent in their current content moderation frameworks.

\noindent\textbf{Dataset scarcity:} Datasets serve as the foundational fuel for generative AI models. However, researchers increasingly face significant obstacles in curating such datasets, primarily due to the high costs of manual annotation and the limitations imposed on large-scale crawling of social media platforms. These challenges have resulted in datasets that are either manually constructed~\cite{kiela2020hateful, 10.1145/3729239} -- often failing to fully capture the richness of human creativity -- or narrowly scoped to specific targets or events~\cite{pramanick2021detecting,pramanick2021momenta,fersini2022semeval}. Moreover, this scarcity of comprehensive datasets has hindered efforts to build robust toxicity detection systems for multimodal social media content. This is in contrast to textual hate speech research, where more extensive taxonomical explorations exist~\cite{sachdeva2022measuring}.

\begin{figure*}[t]
\centering
\includegraphics[
  width=0.95\textwidth,
  trim={10 55 0 30}, 
  clip
]{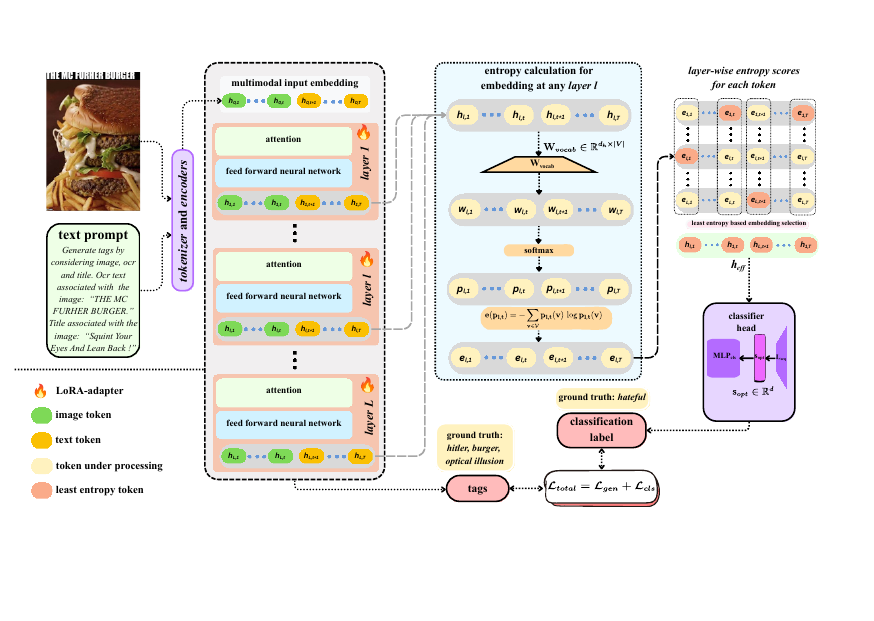}
\caption{\footnotesize Proposed \ours{} framework. Refer to Section~\ref{main:proposed_methodology} for the architectural overview.}
\label{fig:proposed_methodolgy_diagram}
\end{figure*}

\noindent\textbf{Our contributions:} ~\ssubha{\textbf{\textsc{(A)}} We introduce a diverse real-world memes dataset (\datas{}) that incorporates \cam{collaborative tags} collected from associated post metadata-- critical yet often overlooked feature in social media content that is extensively used for describing, categorising, or commenting on digital content\footnote{\url{https://www.imrpress.com/journal/ko/45/6/10.5771/0943-7444-2018-6-500}} . For instance, tags such as \textit{\#hitler, \#jews, \#muslim, \#alabama, \#horse\_meat,} and similar others provide contextual grounding for the corresponding memes.} Unlike prior datasets, it consists of \textit{only} real-world memes with no restrictions based on specific targets or events (see Section~\ref{main:dataset}). The final dataset comprises \textbf{6,300} annotated memes, capturing a wide spectrum of online discourse. 
\textbf{\textsc{(B)}} We design a rigorous two-stage human annotation pipeline. In the first stage, memes are annotated as either \textit{toxic} or \textit{normal}. In the second stage, \textit{toxic} memes are further categorised into one of the three fine-grained classes: \textit{hateful}, \textit{dangerous}, or \textit{offensive}. These categories have been distilled through an iterative annotation process, revealing that a four-class taxonomy is appropriately expressive for moderating a wide range of social media content. This taxonomy can help reduce misclassifications in existing toxicity detection pipelines.
\textbf{\textsc{(C)}} We introduce \ours{}, a novel framework that significantly advances state-of-the-art by generating high-quality \cam{collaborative tags} and improving meme classification performance. The overall architecture of the proposed framework is presented in Figure~\ref{fig:proposed_methodolgy_diagram} and Section~\ref{main:proposed_methodology}.

\begin{tcolorbox}[breakable, colback=gray!5!white,colframe=gray!75!gray,title=Key results]
\footnotesize
\rpoint \ours{} achieves (refer Table~\ref{tab:master_table_comparison}) the best overall performance, outperforming several approaches in the classification task with a Stage I macro-F1 score of \textbf{72.55\%} and a Stage II macro-F1 score of \textbf{64.66\%}; improvements of \textbf{2.07\%} and \textbf{3.12\%}, respectively over the most competing baseline. In addition, it simultaneously generates high-quality \cam{collaborative tags} with scores of \textbf{0.4872} (\sysCHRF{}), \textbf{0.265} (\meteor{}), and \textbf{0.626} (\metSS{}), again outperforming the competing approaches. Our experiments demonstrate that incorporating \cam{collaborative tags} into the multitasking system leads to significant performance gains in meme classification tasks.\\
\rpoint The classification performance of our method \ssubha{on two other popular benchmark hate meme detection datasets compared to four prior approaches demonstrates the generalizability of our method. As presented in Table~\ref{tab:baseline_comparison_toxicity_detection}, \ours{} outperforms the strongest baseline by \textbf{3.30\%} and \textbf{3.26\%} in terms of accuracy and macro-F1 on \fhm{}, and by \textbf{6.47\%} and \textbf{6.25\%} on \mami{}, which shows its effectiveness in various meme classification settings.}\\
\rpoint~\ssubha{Finally, for the tag generation task, we compare our method with three prior approaches. As illustrated in Table~\ref{tab:tag_metrics}, \ours{} achieves improvements of \textbf{0.277} in \sysCHRF{}, \textbf{0.178} in \meteor{}, and \textbf{0.294} in \metSS{}.}
\end{tcolorbox}


\begin{table*}[t]
\scriptsize
\centering
\renewcommand{\arraystretch}{1.5}
\setlength{\tabcolsep}{2pt}

\begin{tabular}{l|c|c|c|c|c|l}
\toprule
\textbf{dataset} &
\makecell[c]{\textbf{event / target}\\ \textbf{independent?}} &
\makecell[c]{\textbf{real world}\\ \textbf{memes?}} &
\makecell[c]{\textbf{post's contextual}\\ \textbf{information?}} &
\makecell[c]{\textbf{two stage}\\ \textbf{annotation?}} &
\textbf{size} &
\textbf{labels} \\
\midrule

\textbf{FHM}~\cite{kiela2020hateful}
& \done & \notdone & \notdone & \notdone
& $\sim$10k
& \makecell[l]{hateful, not-hateful} \\

\textbf{MAMI}~\cite{fersini2022semeval}
& \notdone & \done & \notdone & \notdone
& $\sim$10k
& \makecell[l]{misogynistic, not-misogynistic} \\

\textbf{HARM-C}~\cite{pramanick2021detecting}
& \notdone & \done & \notdone & \notdone
& $\sim$3,544
& \makecell[l]{very/partially harmful, harmless} \\

\textbf{HARM-P}~\cite{pramanick2021momenta}
& \notdone & \done & \notdone & \notdone
& $\sim$3,552
& \makecell[l]{very/partially harmful, harmless} \\

\textbf{UA--RU Conflict}~\cite{thapa2022multi}
& \notdone & \done & \notdone & \notdone
& $\sim$5,680
& \makecell[l]{hateful, not-hateful} \\

\textbf{CrisisHateMM}~\cite{bhandari2023crisishatemm}
& \notdone & \done & \notdone & \notdone
& $\sim$4,723
& \makecell[l]{hateful, not-hateful} \\

\textbf{RUHate-MM}~\cite{thapa2024ruhate}
& \notdone & \done & \notdone & \notdone
& $\sim$20,675
& \makecell[l]{hateful, not-hateful} \\ \hline \hline

\rowcolor[HTML]{E8F1FF}
\textbf{\datas{}}
& \done & \done & \done & \done
& $\sim$6,300
& \makecell[l]{stage \textsc{I}: toxic, normal\\
               stage \textsc{II}: hateful, dangerous,\\
               offensive, normal} \\

\bottomrule
\end{tabular}

\caption{Comparison of \datas{} with existing datasets for toxic meme detection.
Post-contextual information includes metadata such as titles or tags associated with the meme.}
\label{tab:dataset_compare}
\end{table*}

\section{Related works}
\label{main:related_works}

\noindent\textbf{Hate meme detection}: Rapid surge of hate around the globe~\cite{Arcila_Calderon2024-li} in the recent past years, with memes acting as a major source of fuel, has led to the curation of multiple hateful memes datasets~\cite{kiela2020hateful, 10.1145/3729239}. \ssubha{Findings from prior work on MLLMs~\cite{wang-etal-2025-cant} underscore the challenges in robust multimodal safety understanding. Similarly, there has been extensive research in the field to build robust content moderation frameworks~\cite{rizwan2025exploring, mandal2024attentivefusiontransformerbasedapproach, yerukola2025mindgestureevaluatingai, das2023transfer,10.1145/3581783.3613836, 10.1145/3589334.3648145, cao-etal-2022-prompting}, with some works on low-resource languages ~\cite{das2023banglaabusememe,kumari-etal-2024-cm, Xue_Dou_Shi_Li_Gao_2026}. Some works are extended to audio and video modalities~\cite{Boishakhi_2021, 10.1145/3746027.3754553, koushik2026tandemtemporalawareneuraldetection} as well.} Despite such rapid developments, current datasets are either limited to manually curated memes or focus only on a subset of events~\cite{pramanick2021detecting, pramanick2021momenta, chen2023evaluatingexplanationmethodsvisionandlanguage}.

\noindent\textbf{Tags}: \ssubha{Prior research has extensively explored automated tag generation for images~\cite{huang2023tag2text, zhang2024recognize, dai2023exploring}, leading to the development of several dedicated models and datasets. Such synthetic tags have also been used for hate meme detection recently~\cite{koushik2026tandemtemporalawareneuraldetection}. In all the above case, tags refer to general categorizing information used to support resource organization, while \cam{collaborative tagging} refers to the practice of describing, categorizing, or commenting on digital content by users\footnote{\url{https://www.imrpress.com/journal/ko/45/6/10.5771/0943-7444-2018-6-500}, \url{https://www.isko.org/cyclo/tagging}}.} However, to the best of our knowledge, no existing work addresses the specific challenge of tag generation for memes -- visual artefacts that often combine text and imagery in culturally nuanced, context-dependent ways.\\
\noindent\textbf{Present work}: We present a real-world memes dataset (\datas{}) with fine-grained toxicity labels beyond binary classification (Table~\ref{tab:dataset_compare}). \ssubha{In addition, each meme is accompanied by \cam{collaborative tags} obtained from associated post metadata. We also introduce an entropy-based multi-task learning framework \ours{} for open-set image tagging and robust classification.}

\section{The \datas{} dataset}
\label{main:dataset}
\noindent\textbf{Collection:} We source real-world social media memes from [\url{https://imgflip.com}], specifically utilising its \textit{streams} section\footnote{\url{https://imgflip.com/streams}} to crawl meme-centric posts. The platform was chosen for two primary reasons: \textbf{(i)} its strong emphasis on social media-style interactions where memes serve as a central mode of communication, and \textbf{(ii)} its wide variety of user-generated content, allowing us to collect memes without any target-specific or event-specific filtering, thereby closely mirroring real-world social media environments.

For our dataset, we select three high-volume and thematically rich streams---%
\textsc{Dark\_Humour}\footnote{\url{https://imgflip.com/m/Dark_humour}}, \textsc{Memes\_Overload}\footnote{\url{https://imgflip.com/m/MEMES_OVERLOAD}}, and \textsc{Politics}\footnote{\url{https://imgflip.com/m/politics}}---%
based on their popularity and relevance to diverse meme content. We selected these streams through manual inspection of their descriptions and content samples (see Appendix ~\ref{app:platforms} for details). Further, to ensure that the memes included are socially engaging and contextually rich, we manually review a subset of posts and observe that memes with fewer than two comments are often unengaging or irrelevant to a broader audience. Hence, we retain only those memes that received at least two comments. In total, we manually collect 37,072 memes across the selected streams over a period of approximately one month. The memes were then downloaded using a browser-based image downloader extension\footnote{\url{https://chromewebstore.google.com/detail/download-all-images/ifipmflagepipjokmbdecpmjbibjnakm?hl=en}}. As a result, our dataset offers a large and diverse collection of real-world memes, free from artificial filtering or event-driven bias.\\
\noindent\textbf{Metadata}: For each post, along with the meme, we collect its \textbf{(i)} title, \textbf{(ii)} number of comments and \textbf{(iii)} the tags list. For the collected memes, we extract the embedded text using \textsc{Google OCR}\footnote{\url{https://docs.cloud.google.com/vision/docs/ocr}}. To ensure robustness of OCR-extracted text, we perform manual verification of 100 randomly chosen samples and find 98 to be correctly identified, thus depicting the robustness of the tool.\\
\noindent\textbf{Pre-processing}: Before providing the meme along with its metadata for annotation, we perform the following pre-processing steps.\\
\noindent\textbf{(i)} \textit{Deduplication}: Out of the initially collected 37,072 posts, we first apply exact string matching on the \texttt{title} and \texttt{tags} fields to identify duplicate entries. Subsequently, we perform visual deduplication on the matched posts using the \textsc{imagededup}\footnote{\href{https://idealo.github.io/imagededup/methods/hashing/}{imagededup documentation}} library, employing a Hamming distance threshold of zero to ensure strict removal of visually identical images. This two-step deduplication process -- textual followed by visual -- ensures that only unique memes, along with their associated metadata, are retained. After this filtering, we obtain a final dataset comprising 6,300 unique and high-quality memes.\\
\noindent\textbf{(ii)} \textit{Removal of unwanted tags}: To reduce noise and enhance the quality of our tags, commonly occurring irrelevant tags like -- \textit{`darkhumour', `memes', `you have been eternally cursed for reading the tags'} -- are manually removed from each post by expert researchers (refer to Section~\ref{sec:annotation}). Subsequently, we obtain a rich set of 7,209 unique and \cam{collaborative tags} from an initial list of 9,664 unique tags.\\
\noindent\textbf{Agreement}: We strictly adhere to the privacy and copyright regulations of the platform\footnote{\url{https://imgflip.com/terms}} and collect our data only from the publicly available posts. To maintain the privacy of users, we did not store any information that could potentially compromise their anonymity.
\section{Annotation}
\label{sec:annotation}
\begin{figure}[!t]
    \centering
    \includegraphics[width=1\linewidth]{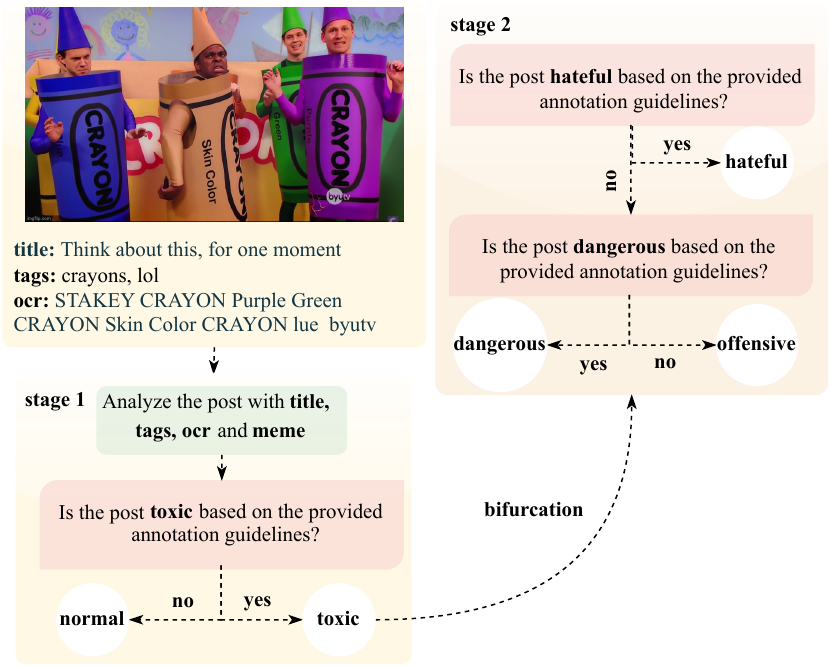}
    \caption{\footnotesize A step-by-step flowchart used by annotators to annotate any post.}
    \label{fig:annotation_flowchart}
\end{figure}
\noindent\textbf{Two-stage annotation}: \ssubha{We follow a two stage annotation process -- \textbf{(i)} classification of each meme into \textit{toxic} or \textit{normal}, and \textbf{(ii)} further bifurcation of toxic memes into \textit{offensive}, \textit{dangerous} or \textit{hateful}. The two-stage process is specifically adopted to mitigate annotation errors as suggested in~\cite{rizwan2025exploring}. We agree that hateful memes can be both offensive and violent, but not vice versa. Hateful memes target a particular community and impose threat and offense upon them. This distinguishes hateful memes from violent and offensive memes. Similary offensive content is limited to abusive or derogatory language, while dangerous content involves expressions that may increase the risk of violence against a someone (for definitions refer to Appendix \ref{subsubsec:definitons}).}\\
 \noindent \ssubha{Prior works also suggest \cite{Mathew_Saha_Yimam_Biemann_Goyal_Mukherjee_2021, Davidson_Warmsley_Macy_Weber_2017} that hate classification is a subjective task, and highlight the importance of distinguishing between closely related but conceptually different forms. Following these prior works, we adopt a mutually exclusive labeling scheme, where each instance is assigned a single primary category. This design is also consistent with real-world content moderation systems, where most streaming and social media platforms (e.g., YouTube\footnote{\url{https://support.google.com/youtube/answer/9288567}}, Facebook\footnote{\url{https://www.facebook.com/privacy/center}}) typically allow users to select only one primary violation category. Figure~\ref{fig:social_media_policies} in the Appendix presents snapshots of the policy/reporting interfaces of Facebook, Youtube, and LinkedIn, where users are generally guided to select a single primary violation category.\\} 
\ssubha{
\noindent In subjective tasks such as hate speech or toxicity detection, annotator agreement alone does not necessarily guarantee validity due to variability in lifestyle, culture, and demographic backgrounds. Prior work~\cite{rottger-etal-2022-two} distinguishes between the prescriptive paradigm, which enforces a consistent labeling standard, and the descriptive paradigm, which captures diverse annotator beliefs. In this work, we adopt a \textbf{prescriptive annotation paradigm} to ensure consistency and reduce ambiguity in labeling. During the annotation process, annotators were instructed to strictly adhere to the label definitions.}\\
Figure~\ref{fig:annotation_flowchart} presents a brief pipeline of the employed annotation process over two stages. The instructions that we designed are given below, and some samples from those provided to annotators are present in Table~\ref{tab:annotation_samples} in the Appendix.  Further, we also conducted a pilot study at each stage before progressing to the final annotations. For both stages, we perform \textbf{three annotations} per sample by three different annotators to minimise annotation bias. We use the standard definition of labels that are effectively used for practical applications. In the upcoming subsections, we discuss each stage separately and present further detailed annotation instructions with corresponding definitions of the labels in Appendix~\ref{app:annotation_guidelines}.

\begin{figure*}[!htbp]
    \centering
    \includegraphics[width=1\linewidth]{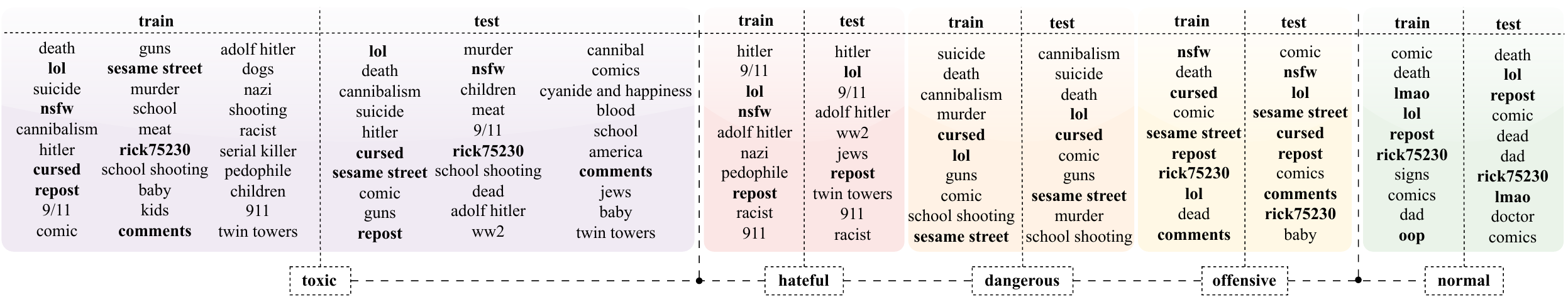}
    \caption{\footnotesize Frequently occurring tags in the \textit{toxic}, \textit{hateful}, \textit{offensive}, and \textit{normal} classes.}
    \label{fig:top_tags}
\end{figure*}

\noindent\textbf{Annotator details}: \ssubha{To ensure a high-quality annotation process, we shortlisted 25 annotators based on the following minimum criteria: \textbf{(i)} candidates having professional experience and prior work aligned with our research, and \textbf{(ii)} familiarity with diverse social media content. The pool included 14 male and 11 female participants. To ensure high-quality annotations, we selected 19 Master's students and 6 PhD students aged between 22 and 40 years. In addition to the above, the group included diverse religious backgrounds: 14 identified as Hindus, 8 as Muslims, and 3 as Christians. We additionally filtered annotators by asking them to pass a qualification task curated by the supervising researchers to assess label comprehension and consistency. The entire annotation process was conducted under the close supervision of two PhD researchers. We detail the safety guidelines in Appendix~\ref{app:annotation_guidelines}; further, as an extra safety measure, we rolled out a maximum of only 50 samples per day to each annotator.}\\
\noindent\textbf{Annotation codebook}: We also provide detailed annotation instructions, accompanied by 25 samples that have been manually selected and annotated by the two supervising researchers. As outlined in Appendix~\ref{app:annotation_guidelines}, our annotation codebook has been constructed based on standard guidelines from which we have derived our definitions. For the toxic label, we strictly follow the definition used by the Perspective API, given its widespread adoption in production settings. For the hateful label, we adhere to the Facebook hate meme definition~\cite{kiela2020hateful}, which has become the de facto standard. For the dangerous label, we follow the annotation guidelines available at the following source\footnote{\href{https://www.dangerousspeech.org/dangerous-speech}{https://www.dangerousspeech.org/dangerous-speech}}.
\subsection{Stage I: Binary labelling of the dataset}
\noindent\textbf{Annotation}: The full set of 6,300 samples has been evenly distributed among the 12 selected annotators (out of the 25 recruited). As stated earlier, we employ three annotators per sample, which enhances annotation diversity. We achieve a final inter-annotator agreement score of  0.753, as measured by Fleiss' $\kappa$. Notably, for a subjective task~\cite{Mathew_Saha_Yimam_Biemann_Goyal_Mukherjee_2021} such as ours, this level of agreement is considered relatively high, thereby validating the quality of our annotations. Each annotator has been paid USD 40 for this task, which is much above the minimum wage in the annotators' country.\\
\noindent\textbf{Label assignment}: The final label for each sample is determined based on the majority vote of the three annotations, i.e., the label with at least two agreements is considered to be the final label. In this stage, we finally obtained 1,446 normal and 4,854 toxic samples (refer to Table~\ref{tab:dataset_statistics} for complete statistics).

\subsection{Stage II: Fine-grained labelling}
\noindent\textbf{Annotation}: As a pilot step, we randomly sampled 250 toxic memes from the previous stage and had the expert annotators categorise them into three distinct labels. These labelled samples were then evenly distributed among the 12 annotators, each of whom annotated all the assigned samples. Upon verification, we found that the annotators were performing satisfactorily and thus proceeded with the final annotation with this group. Subsequently, the set of 4,854 toxic samples was categorised into three classes: \textit{hateful}, \textit{dangerous}, and \textit{offensive}. As before, each sample received three independent annotations. The annotators achieved a Fleiss’ $\kappa$ agreement score of 0.793 -- a slight improvement over Stage I -- highlighting the value of multi-stage and multi-pilot studies in enhancing annotation quality.\\
\noindent\textbf{Label assignment}: The final label is determined based on the majority vote of the annotators for each sample. \ssubha{In total, 26 samples were identified as undecidable cases, where majority voting failed. These undecidable instances mainly include posts containing mixed linguistic cues, subtle contextual dependencies or the coexistence of multiple possible categories. Given the relatively small number of such instances, final annotations for these cases were determined by expert researchers following the same annotation guidelines to maintain consistency.} The overall dataset statistics are noted in Table~\ref{tab:dataset_statistics}. To maintain the diversity of tags in both the splits, we ensure that each tag appears at least 15\% of the time in the test split, resulting in a total of 1,000 test samples and 5,300 training samples. Word clouds for label-wise tags are presented in Table~\ref{fig:word_clouds} of Appendix~\ref{sec:more_analysis}.

\begin{table}[t]
\scriptsize
\centering
\renewcommand{\arraystretch}{1.2}
\setlength{\tabcolsep}{2mm}
\begin{tabular}{c c | c c | c}
\hline
\rowcolor[HTML]{FFF2CC}
\textbf{stage} & \textbf{label} & \textbf{train} & \textbf{test} & \textbf{total} \\
\hline
\textbf{I \& II} & normal    & 1297 & 149 & 1446 \\
\hline
\textbf{I}       & toxic     & 4003 & 851 & 4854 \\
\hline
\multirow{4}{*}{\textbf{II}} 
                 & hateful   & 1470 & 282 & 1752 \\
                 & dangerous & 1847 & 472 & 2319 \\
                 & offensive & 686  & 97  & 783  \\
\hline
\multicolumn{2}{c|}{\textbf{total}} & 5300 & 1000 & 6300 \\
\hline
\end{tabular}
\caption{\footnotesize Dataset statistics showing the distribution of samples across classes.}
\label{tab:dataset_statistics}
\end{table}


\begin{table}[!ht]
\scriptsize
\centering
\renewcommand{\arraystretch}{1}
\begin{tabular}{p{4.5cm}|p{2cm}}
\rowcolor[HTML]{FFF2CC}
\multicolumn{1}{c|}{\cellcolor[HTML]{FFF2CC}\textbf{tag pairs}}                                             & \textbf{theme}                         \\ \hline
(`hitler', `nazi'), (`hitler', `jews'), (`hitler', 'ww2'), (`hitler', `holocaust'), (`jews', `nazi'), (`concentration camp', `nazi'), (`hitler', `stalin') &
  Historical events / World War II \\ \hline
(\texttt{`9/11'}, `twin tower'), (`911', `twin tower'), (\texttt{`9/11'}, 'muslim'), (`cow', `muslim'), (`911 \texttt{9/11} twin tower impact', `muslim') &
  \texttt{9/11} and Islamophobic references \\ \hline
(`jesus', `satan'), (`satan', `the bible'), (`jesus', `the bible')                                          & Religious contrast / satire            \\ \hline
(`atomic bomb', `hiroshima'), (`hiroshima', `ww2'), (`hiroshima', `nuke')                                   & Nuclear warfare / World War II         \\ \hline
(`alabama', `incest'), (`incest', `sweet home alabama'), (`alabama', `dad')                                 & Southern US stereotypes, taboo, humour \\ \hline
(`cannibal', `cannibalism'), (`fresh', `meat'), (`cannibalism', `meat'), (`human', `meat')                  & Creepy / disturbing themes             \\ \hline
(`mass shooting', `school shooting'), (`gun', `school shooting'), (`gun', `school'), (`gun', `usa'), (`mass shooting', `usa'), (`mass shooting', `america') &
  US gun violence \\ \hline
(`baby', `dead'), (`baby', `cannibalism')                                                                   & Child-related violence / abortion      \\ \hline
(`africa', `water'), (`africa', `hungry'), (`africa', `starve')                                             & Humanitarian crises in Africa          \\ \hline
(`black people', `racist'), (`black', `black life matter'), (`black', `white'), (`angry black guy', `lame') & Racism                                 \\ \hline
(`elmo', `sesame street'), (`big bird', `sesame street'), (`ernie', `sesame street'), (`mayor', `serial killer'), (`free candy van', `sonic the hedgehog') &
  Cartoon / anime characters in dark contexts \\ \hline
(`world war 3', `ww3'), (`ukraine', `ww3'), (`gaza', `israel'), (`monster', `ww3')                                                                                      & Global conflict / war escalation  \\
\hline
\end{tabular}
\caption{\footnotesize Top co-occurring tag pairs and associated semantic themes.}
\label{tab:cooccur}
\end{table}

\section{Analysis of the dataset}
\label{main:analysis}

One of the most unique features of our dataset is the tags associated with each meme. This allows us to perform certain interesting analyses that we report below. \\
\noindent \textbf{Frequent tags:} Some of the most frequent tags we find in the train and the test splits are illustrated in Figure~\ref{fig:top_tags}. We report the top 30 tags from the toxicity class as well as the top 10 tags each from the hateful, dangerous, offensive and normal classes, respectively. The key observations are as follows.\\ 
\noindent\textbf{(i)} The distribution of the top tags across the train and the test splits is very similar, indicating that our strategy of splitting the data is effective.\\
\noindent\textbf{(ii)} The most prevalent tags in the \textit{hateful} class are \textit{`9/11', `nazi', `adolf hitler', `paedophile'} and \textit{`racist'}, indicating their popularity in spreading hateful sentiments. We also observe that for the \textit{dangerous} class, some tags that appear benign at the surface level and are primarily associated with children's media show up. These include \textit{`ernie and bert', `sesame street'}, and \textit{`sonic the hedgehog'}, which are used to render a comic angle to the dangerous posts (see Table~\ref{tab:example_memes_analysis} in the Appendix for more examples).\\
\noindent\textbf{(iii)} We observe that users are generally reluctant to reshare content involving serious violence, suicide, or murder, as indicated by the relatively low frequency of the \textit{`repost'} tag in the \textit{dangerous} category compared to its prevalence in other categories.\\
\noindent\textbf{(iv)} The widespread use of the \textit{`lol'} tag suggests an attempt to downplay harmful content through humour.\\
\noindent\textbf{(v)} Frequent use of tags such as \textit{'cannibalism', 'murder', 'suicide'}, and \textit{'abortion'} and violent phrases like \textit{`school shooting'} underscores how dark humour is often employed as a tool to normalise or propagate digital violence, thereby contributing to a heightened sense of danger.\\
\noindent\textbf{(vi)} One notable observation concerns the use of the tag \textit{`alabama'}, which frequently appears in hateful memes that mock familial relationships, highlighting how geographic stereotypes are weaponised for humour and hate.\\
\noindent\textbf{Related tags}: To understand the semantic themes associated with frequently occurring tag pairs, we conduct co-occurrence analysis. Prior to the analysis, all tags are lemmatised to ensure consistency and improve result accuracy. Our primary objective is to identify how often specific tags co-occur and what thematic contexts they represent within the meme content. Some of the top co-occurring pairs and their themes are listed in Table~\ref{tab:cooccur}. \ssubha{The co-occurring tag pairs in the table indicate that toxic meme content is often structured around semantically linked concepts. For example, themes such as \textit{Historical events / World War II} frequently associate tags like \textit{`hitler'} and \textit{`nazi'}. Similarly, in the context of \textit{\texttt{9/11} and Islamophobic references}, tags such as \textit{`twin tower', `9/11'}, and \textit{`muslim'} often co-occur. Further, tags like \textit{`africa'} and \textit{`water'} are grouped under themes related to \textit{Humanitarian crises in Africa}. These patterns suggest that toxicity is reinforced through well-established contextual associations. The theme of \textit{Cartoon / anime characters in dark contexts} indicates that such characters often co-occur with disturbing or violent concepts, suggesting that they are used to amplify humour or shock value. Overall, this analysis highlights that meme semantics inherently arises from the interaction of the multiple \cam{collaborative tags.}}

\section{The \ours{} framework}
\label{main:proposed_methodology}
\subsection{Entropy-guided layer selection objective}
\label{main:entropy_guided_framework}

\ssubha{Past approaches~\cite{cao-etal-2022-prompting, cao2023pro, cao2024modularizednetworksfewshothateful, hee2024bridging, lu2025having} in toxic meme detection have focused primarily on single-task setups and rely either on zero-/few-shot prompting or perform low-resource fine-tuning using strategies like PEFT/LoRA.} Recently, multimodal research has evolved, and some works like~\cite{he2025seeing,luo2024vision} propose a deeper understanding of the model's internal representations and semantic alignment. Recent empirical evidence presented in~\cite{chen-etal-2025-multimodal} suggests that deep representational alignment between vision and language modalities peaks at intermediate depths before undergoing representational drift in terminal layers. Taking cues from these works, we propose a novel token confidence-based approach for simultaneous tag generation and toxic meme classification. Below, we detail our approach. \\
Figure~\ref{fig:proposed_methodolgy_diagram} illustrates the complete workflow of \ours{}. We consider a multimodal model \(\mathcal{M}\) that takes a text prompt \(\mathcal{T}\) and an image \(\mathcal{I}\) as input and consists of \(L\) stacked layers. Given an input sequence of fixed length, the model produces hidden representations across layers. Let \(l \in \{1,\dots,L\}\) \cam{(including the token embedding layer) denote the layer index, \(t \in \{1,\dots,T\}\)} denote the token position in the input sequence, and \(v \in V\) denote a token from the model vocabulary \(V\). The hidden state \(\mathbf{h}_{l,t} \in \mathbb{R}^{d_h}\) corresponds to token \(t\) at layer \(l\), as depicted in the figure, where \(d_h\) represents the hidden dimension of the model. To process the model’s internal state, we extract token embeddings from all layers and project them into the model’s vocabulary space using the pretrained language model head \(\mathbf{W}_{\text{vocab}} \in \mathbb{R}^{d_h \times |V|}\). This projection produces \(\mathbf{w}_{l,t}\), which indicates the representation of the \(t\)-th token at layer \(l\) in the vocabulary V:
\begin{equation}
\mathbf{w}_{l,t} = \mathbf{h}_{l,t}\,\mathbf{W}_{\text{vocab}}.
\end{equation}

\noindent A softmax function is applied to \(\mathbf{w}_{l,t}\) to obtain the confidence distribution \(p_{l,t}\) over the vocabulary for the \(t\)-th token at layer \(l\):
\begin{equation}
p_{l,t} = \mathrm{softmax}\!\left(\mathbf{w}_{l,t}\right),
\end{equation}

\noindent To measure the confidence of the representation at layer \(l\) for token \(t\), we compute the Shannon entropy~\cite{vajapeyam2014understandingshannonsentropymetric} \(e_{l,t}\), which represents the entropy of token \(t\) at layer \(l\), as follows:
\begin{equation}
\label{eq:entropy_calc}
e_{l,t} = - \sum_{v \in V} p_{l,t}(v)\,\log p_{l,t}(v).
\end{equation}
This entropy measure enables the dynamic selection of the least random token embedding across layers by selecting the representation corresponding to the minimum entropy, enabling the classification head to extract features from the most certain internal representations \(\mathbf{h}_{\text{eff}}\):
\begin{equation}
\mathbf{h}_{\text{eff}} =
\mathrm{Aggregate}\!\left(
\left\{
\mathbf{h}_{l^{*}
,t}
\;\middle|\;
l^{*} = \arg\min_{l \in \mathcal{L}} e_{l,t}
\right\}_{t=1}^{T}
\right).
\end{equation}

\noindent Finally, the confident sequence representation \(\mathbf{h}_{\text{eff}}\) is passed through the classification head to predict the class logits. The classification head consists of a sequence compression layer that compresses the sequence and projects it to a fixed-dimensional representation, which is then passed through an Multi-Layer Perceptron(MLP) for final classification:
\begin{equation}
\textrm{classification label} = \mathrm{MLP}\!\left(\mathrm{Compress}\!\left(\mathbf{h}_{\text{eff}}\right)\right)
\end{equation}
\noindent \ssubha{We employ a weighted cross-entropy loss for the classification task, which is depicted as \(\mathcal{L}_{\text{cls}}\) in the figure. To address class imbalance, class weights are computed following the weighting scheme described in~\cite{3709}\footnote{\href{https://scikit-learn.org/stable/modules/generated/sklearn.utils.class_weight.compute_class_weight.html}{scikit-learn documentation}}}.

\begin{table*}[!t]
\centering
\footnotesize
\renewcommand{\arraystretch}{1.1}
\setlength{\tabcolsep}{3.5pt}
\colorlet{groupbg}{gray!10}
\colorlet{bestbg}{green!12}

\begin{tabular}{
  l l l
  S[table-format=2.2]
  S[table-format=2.2]
  S[table-format=1.4]
  S[table-format=1.4]
  S[table-format=1.4]
}
\toprule
\multirow{2}{*}{\textbf{approach}}
  & \multirow{2}{*}{\textbf{model}}
  & \multirow{2}{*}{\textbf{shot}}
  & \multicolumn{2}{c}{\textbf{mF1$\uparrow$}}
  & \multicolumn{3}{c}{\textbf{tag similarity$\uparrow$}} \\
\cmidrule(lr){4-5}\cmidrule(lr){6-8}
  & & & \textbf{\textsc{stage I}} & \textbf{\textsc{stage II}}
      & \textbf{\textsc{chrF}} & \textbf{MET} & \textbf{SS} \\
\midrule

\rowcolor{groupbg}
\multicolumn{8}{l}{\textbf{multimodal prompting}} \\

zero-shot

& \sysGPT{} & 0 & 54.57 & 41.62 & {--} & {--} & {--} \\
& \sysLV{}  & 0 & 45.84 & 10.19 & {--} & {--} & {--} \\[1pt]
\multicolumn{8}{c}{\dotfill} \\
\multirow{4}{*}{\begin{tabular}[c]{@{}l@{}}
few-shot\\
\textit{(random)}
\end{tabular}}
& \multirow{2}{*}{\sysGPT{}}
& 2 & 61.09 & 43.44 & {--} & {--} & {--} \\
& & 4 & 62.46 & 44.07 & {--} & {--} & {--} \\

& \multirow{2}{*}{\sysLV{}}
& 2 & 49.28 & 21.32 & {--} & {--} & {--} \\
& & 4 & 51.50 & 24.26 & {--} & {--} & {--} \\[1pt]
\multicolumn{8}{c}{\dotfill} \\
\multirow{4}{*}{\begin{tabular}[c]{@{}l@{}}
few-shot\\
\textit{(tag similarity)}
\end{tabular}}
& \multirow{2}{*}{\sysGPT{}}
& 2 & 63.96 & 50.48 & {--} & {--} & {--} \\
& & 4 & 64.86 & 52.88 & {--} & {--} & {--} \\

& \multirow{2}{*}{\sysLV{}}
& 2 & 59.99 & 40.62 & {--} & {--} & {--} \\
& & 4 & 59.71 & 41.48 & {--} & {--} & {--} \\[1pt]
\multicolumn{8}{c}{\dotfill} \\
\multirow{4}{*}{\begin{tabular}[c]{@{}l@{}}
few-shot\\
\textit{(image similarity)}
\end{tabular}}
& \multirow{2}{*}{\sysGPT{}}
& 2 & 61.86 & 48.59 & {--} & {--} & {--} \\
& & 4 & 64.89 & 50.98 & {--} & {--} & {--} \\

& \multirow{2}{*}{\sysLV{}}
& 2 & 54.62 & 34.57 & {--} & {--} & {--} \\
& & 4 & 55.05 & 35.66 & {--} & {--} & {--} \\

\midrule
\rowcolor{groupbg}
\multicolumn{8}{l}{\textbf{unimodal prompting}} \\
zero-shot
        & \sysGPT{}    & {0} & 46.72 & 30.22 & {--} & {--} & {--} \\
          & \textsc{Llama}   & {0} & 48.07 & 27.17 & {--} & {--} & {--} \\
          & \textsc{Mistral} & {0} & 55.48 & 31.60 & {--} & {--} & {--} \\

\midrule
\rowcolor{groupbg}
\multicolumn{8}{l}{\textbf{baselines}} \\
PH~\cite{cao-etal-2022-prompting}
          & {--} & {--} & 61.83 & {--}  & {--} & {--} & {--} \\
PC~\cite{cao2023pro}
          & {--} & {--} & 63.26 & {--}  & {--} & {--} & {--} \\
MH~\cite{cao2024modularizednetworksfewshothateful}
          & {--} & {--} & 49.84 & {--}  & {--} & {--} & {--} \\
BM~\cite{hee2024bridging}
          & {--} & {--} & 59.70 & {--}  & {--} & {--} & {--} \\

\midrule
\rowcolor{groupbg}
\multicolumn{8}{l}{\textbf{ablations: last hidden layer}} \\
uni-task  & \sysLV{}  & {--} & 61.58 & 44.69 & {--}   & {--}   & {--}   \\
multi-task& \sysLV{}  & {--} & 65.03 & 54.56 & 0.2585 & 0.1050 & 0.4540 \\
\multicolumn{8}{c}{\dotfill} \\

uni-task  & \sysP{}   & {--} & 61.52 & 45.92 & {--}   & {--}   & {--}  \\
multi-task& \sysP{}   & {--} & 70.48 & 61.54 & 0.4782 & 0.2600 & 0.6220   \\

\midrule
\rowcolor{groupbg}
\multicolumn{8}{l}{\textbf{proposed: \ours{}}} \\

uni-task       & \sysLV{} & {--} & 62.19 & 51.90 & {--}   & {--}   & {--}   \\
multi-task     & \sysLV{} & {--} & 67.24 & 57.99 & 0.2648 & 0.1070 & 0.4470 \\
\multicolumn{8}{c}{\dotfill} \\
uni-task       & \sysP{}  & {--} & 60.85 & 42.78 & {--}   & {--}   & {--}   \\
\rowcolor{bestbg}
multi-task     & \sysP{}  & {--} & \textbf{72.55} & \textbf{64.66}
                             & \textbf{0.4872} & \textbf{0.2650} & \textbf{0.6260} \\
\bottomrule
\end{tabular}

\caption{
  Comparative results for toxicity detection and tag similarity.
  Best results are
  \textbf{Abbrev.:}
  PH: \textsc{PromptHate}; PC: \textsc{ProCap}; MH: \textsc{ModHate}; BM: \textsc{Bridging Modalities}; mF1: Macro-F1; MET: \meteor{}; SS: \metSS{}; \,{--}\, denotes \, unsupported tasks.\,
}
\label{tab:master_table_comparison}
\end{table*}

\subsection{Multitask learning objective}
\label{main:multi_tasking_framework}
\noindent A meme might not always be accompanied by tags, unlike in the case of our dataset. It is therefore important to have an automatic method to assign tags to an arbitrary input meme, which could enhance toxicity detection. As a combined objective, we therefore propose a novel multitasking framework where we jointly train toxic tag generation and discriminative label classification. The main motivation of this framework is to leverage semantic supervision to guide discriminative classification, where the embeddings are shaped into a representation space where the supervision provided by the generative task acts as a regulariser for the shared representation space. Our experimental observations (refer to Section~\ref{main:results}) reveal that supervising the model to generate explicit toxic tags stimulates the intermediate hidden representations of the model and boosts discriminative performance, thereby signalling that the model learns more detailed and meaningful representations in joint training. The overall loss is calculated as a combination of the generation and classification losses:
\begin{equation}\label{eq:total_loss}
\mathcal{L}_{total} = \mathcal{L}_{gen} + \mathcal{L}_{cls} 
\end{equation}
\noindent where, $\mathcal{L}_{gen}$ represents the generative loss and $\mathcal{L}_{cls}$ represents the classification loss. This alignment makes the layer selection objective described in Section~\ref{main:entropy_guided_framework} more robust as a toxicity classifier. \ssubha{Specifically, $\mathcal{L}_{gen}$ is the default token-level cross-entropy loss over the generated tag sequence, whereas $\mathcal{L}_{cls}$ is defined as a weighted cross-entropy loss for the classification task. }The details on the employed models and experimental setup are elaborated in Appendix~\ref{app:models} and~\ref{app:experimental_setup}, respectively. In contrast, the uni-tasking framework (see Table~\ref{tab:master_table_comparison}), considers only the classification loss $\mathcal{L}_{\text{cls}}$, since it is solely dedicated to toxicity detection. In the next section, we detail our empirical observations.


\section{Results}
\label{main:results}

In this section we assess the performance of \ours{} toxic meme detection on the test split of the \datas{} dataset for both stages -- I and II. We also present the results of tag generation performance achieved by \ours{}. For our experiments, we use two open-source VLMs -- \sysP{} and \sysLV{} (refer Appendix~\ref{app:models} 
and~\ref{app:experimental_setup} for further details). Table \ref{tab:master_table_comparison} shows the experimental results. \\
\noindent\textbf{Toxic meme detection}: \ours{} with the multi-task learning objective, it consistently outperforms uni-task training across both models and stages, highlighting the benefit of joint learning for toxicity detection and tag generation, as summarised in Table~\ref{tab:master_table_comparison}. \sysP{} attains the highest macro F1 scores of 72.55 and 64.66 for stages I and II, respectively. This represents a notable improvement over the uni-task setting, which has macro F1 scores of 60.85 in stage I and 42.78 in stage II. \ssubha{A similar pattern is also observed for \sysLV{}, where \ours{} in the multi-task framework achieves macro F1 scores of 67.24 and 57.99 for the two stages, respectively, outperforming the uni-task learning approach. Likewise, \ours{} also outperforms the setup that uses the last hidden layer to compute $\mathcal{L}_{cls}$ instead of the proposed entropy-based scheme. \\
\noindent\textit{Zero-shot and few-shot variants}: Notably, \ours{} outperforms several zero-shot and few-shot variants illustrated in Table~\ref{tab:master_table_comparison}. Below, we provide a brief overview of these methods (for prompts, refer to Appendix~\ref{main:prompt}).} \ssubha{To assess the pretrained capabilities of the popular VLM models, we perform zero-shot evaluation using \sysLV{} and the powerful \sysGPT{} by giving input as an image, text and title associated with the image. Although \sysP{} is used as the backbone of our framework, it has been red-teamed and cannot be reliably used for hate classification without task-specific fine-tuning. We also investigate whether the tags associated with a meme are themselves sufficent for prediction (i.e., the meme image is not part to the input). To test this, we consider pretrained LLM models -- \textsc{Mistral}, \textsc{Llama} and \sysGPT{} and perform the prediction by providing a prompt having only the tags associated with the meme as input.}\\
\noindent\ssubha{We further investigate, under consistent model settings, few-shot variants with diverse sampling strategies to ensure a fair comparison. Few-shot sampling strategies are as follows. \text{(i)} \textit{Few shot with random sampling}: In this setting, samples are chosen randomly with different seed values such as 19, 42, and 97. \ssubha{\text{(ii)} \textit{Few shot with image similarity}: Instead of selecting random shots, we
select those memes as few-shot examples that are most similar to the query meme in terms of the similarity between
their CLIP-based image embeddings. \ours{} outperforms all these zero-/few-shot variants as observed in Table~\ref{tab:master_table_comparison}.} \text{(iii)} \textit{Few shot with tag similarity}: The few-shot examples are selected based on the ground-truth tag similarity of the memes. To obtain these few-shot samples, we first compute the CLIP embeddings for each tag in the train set and also those that are associated with the query sample. Next, we find the maximum cosine similarity of each query tag embedding with the tag embeddings of an example meme from the train set. For instance, given a test sample with tags $\{t_1, t_2\}$ and an example meme with tags $\{e_1, e_2, e_3\}$, we compute the cosine similarity between the CLIP embedding of tag $t_1$ and those of $e_1$, $e_2$, and $e_3$, and record the maximum similarity value. We repeat the same process for $t_2$. The final tag similarity between the test and the example sample is then defined as the mean of the maximum similarity scores. }\\
\ssubha{\noindent\textit{Competing baselines and generalizability across datasets}: To further ground \ours{}, we compare it with the existing SOTA methods. We compare these methods using \datas{} as well as two well-studied benchmark datasets (\fhm{} and \mami{}). The methods that we choose are \textsc{PromptHate}~\cite{cao-etal-2022-prompting}, \textsc{ProCap}~\cite{cao2023pro}, \textsc{ModHate}~\cite{cao2024modularizednetworksfewshothateful} and \textsc{Bridging Modalities}~\cite{hee2024bridging}. \\
Table~\ref{tab:master_table_comparison} shows that \ours{} demonstrates substantial improvement over all these existing approaches for the \datas{} dataset. In case of \fhm{} and \mami{} datasets as benchmarks, we do not have tags that is necessary to train \ours{}. For the training points of both these datasets we therefore infer the tags using a \sysP{} model that is fine-tuned on the \datas{} dataset to predict tags. Next, we use the gold labels and these inferred silver tags to train \ours{} (\sysP{} backbone, as it exhibits the best performance in  Table \ref{tab:master_table_comparison}) for each of the two datasets. Table \ref{tab:baseline_comparison_toxicity_detection} compares the results of the baselines with \ours{}. Our approach shows substantial improvement over the other baselines for the \fhm{} and \mami{} datasets, achieving a macro F1 score of 78.11 and 79.67 and an accuracy of 78.4 and 80.1, respectively. In contrast, the best-performing baseline (\textsc{ProCap}) achieves a maximum accuracy and macro F1 score of 75.10 and 74.85 on \fhm{}, and 73.63 and 73.42 on \mami{}.}
\begin{table}[t]
\centering
\scriptsize
\renewcommand{\arraystretch}{1.2}
\setlength{\tabcolsep}{4pt}
\begin{tabular}{lcccc}
\toprule
\multirow{2}{*}{\textbf{Model}} 
& \multicolumn{2}{c}{\textbf{FHM}} 
& \multicolumn{2}{c}{\textbf{MAMI}} \\
\cmidrule(lr){2-3} \cmidrule(lr){4-5}
& \textbf{acc} & \textbf{mF1} 
& \textbf{acc} & \textbf{mF1} \\
\midrule
PH   & 72.98 & 72.24 & 70.31 & 70.18 \\
PC   & 75.10 & 74.85 & 73.63 & 73.42 \\
MH   & 57.60 & 53.88 & 69.05 & 68.78 \\
BM   & 66.00 & 65.80 & 70.50 & 70.10 \\
\midrule
\ours{} & \colorbox{green!20}{\textbf{78.4}} & \colorbox{green!20}{\textbf{78.11}} & \colorbox{green!20}{\textbf{80.1}} & \colorbox{green!20}{\textbf{79.67}} \\
\bottomrule
\end{tabular}
\caption{Performance comparison of our method with baseline approaches on the widely used benchmark datasets FHM and MAMI.  Best results are \colorbox{green!20}{\textbf{highlighted.}} Our model outperforms baseline methods in terms of accuracy (Acc) and macro-F1 (mF1) across both datasets. PH: \textsc{PromptHate}, PC: \textsc{ProCap}, MH: \textsc{ModHate}, BM: \textsc{Bridging Modalities}.}
\label{tab:baseline_comparison_toxicity_detection}
\end{table}



\begin{table}[!ht]

\centering
\scriptsize
\renewcommand{\arraystretch}{1.2}
\setlength{\tabcolsep}{3mm}
\begin{tabular}{l|ccc}
\hline
\multirow{2}{*}{\textbf{model}} & \multicolumn{3}{c}{\textbf{tag similarity}} \\ \cline{2-4} 
                                   & \textbf{chrF} & \textbf{met} & \textbf{SS} \\ \hline

RAM++              & 0.1942 & 0.058 & 0.406 \\
RAM++ (O)                     & 0.094 & 0.012 & 0.279 \\
RAM                   & 0.19 & 0.05 & 0.40 \\
RAM (P)                  & 0.087 & 0.014 & 0.338 \\
RAM (O)                       & 0.079 & 0.013  & 0.285 \\
\textsc{Tag2Text}    & 0.1631 & 0.041 & 0.372 \\
\textsc{Tag2Text} (P)     & 0.2102 & 0.087 & 0.332 \\
\hline
\ours{} & \colorbox{green!20}{\textbf{0.4872}} & \colorbox{green!20}{\textbf{0.265}} & \colorbox{green!20}{\textbf{0.626}} \\
 \hline
\end{tabular}
\caption{\footnotesize Performance of the tag generation module. Best results are \colorbox{green!20}{\textbf{highlighted.}} 
chrF: \sysCHRF{}, met: \meteor{}, SS: \metSS{}, 
(O): open-set, (P): pre-trained using \datas{}.}
\label{tab:tag_metrics}
\end{table}

    
    

\noindent\textbf{Tag generation module}: As stated earlier, Table~\ref{tab:master_table_comparison} presents the tag generation results over \datas{} dataset as well. The metrics used are \sysCHRF{}{}, \meteor{} and \metSS{}\footnote{\url{https://huggingface.co/tasks/sentence-similarity}}. For each ground-truth tag, we compute each of the above metrics with all the generated tags and take the maximum value among these. Next, we average this maximum over all the ground-truth tags and calculate mean of these averages over all the test data points. Among the two models used within the \ours{} multi-tasking framework, \sysP{} consistently outperforms \sysLV{} in terms of \sysCHRF{}{}, \meteor{} and \metSS{}. Specifically, \sysP{} achieves scores of 0.4872 (\sysCHRF{}{}), 0.265 (\sysCHRF{}{}) and 0.626 (\metSS{}), whereas \sysLV{} attains 0.2648, 0.107 and 0.447, respectively. These results indicate that \sysP{} is more effective than \sysLV{} for tag generation in the multi-tasking framework \ours{}.\\
\noindent\textit{Tag generation baselines}: Since \ours{} generate tags along with toxicity labels, it is imperative to compare its tag generation performance with state-of-the-art generators. To this purpose, we use three state-of-the-art baselines for evaluation namely \textsc{Tag2Text}~\cite{huang2023tag2text}, \textsc{RAM}~\cite{zhang2024recognize}, and \textsc{RAM++}~\cite{10.1145/3746027.3755316}. To ensure fair evaluation, we further pre-trained the \textsc{Tag2Text} and \textsc{RAM} models on our \datas{} dataset, allowing them to suitably adapt to domain-specific toxic vocabulary. Table~\ref{tab:tag_metrics} shows that \ours{} by far outperforms all the other tag generation baselines in terms of all the three metrics -- \sysCHRF{}, \meteor{} and \metSS{}. State-of-the-art models fail to capture the meme's social context and their implicit meaning. They focus primarily on the visible objects in the image (refer to Table~\ref{fig:model_compare_gen_tags}; also refer to Table~\ref{tab:model_compare_gen_tags_extra_examples} in Appendix for more examples).

\begin{table}[!ht]
\scriptsize
\centering
\setlength{\tabcolsep}{0.1pt}
\begin{tabular}{|c|c|c|}
\hline
\textbf{Image} & \includegraphics[width=0.3\columnwidth,height=2.5cm]{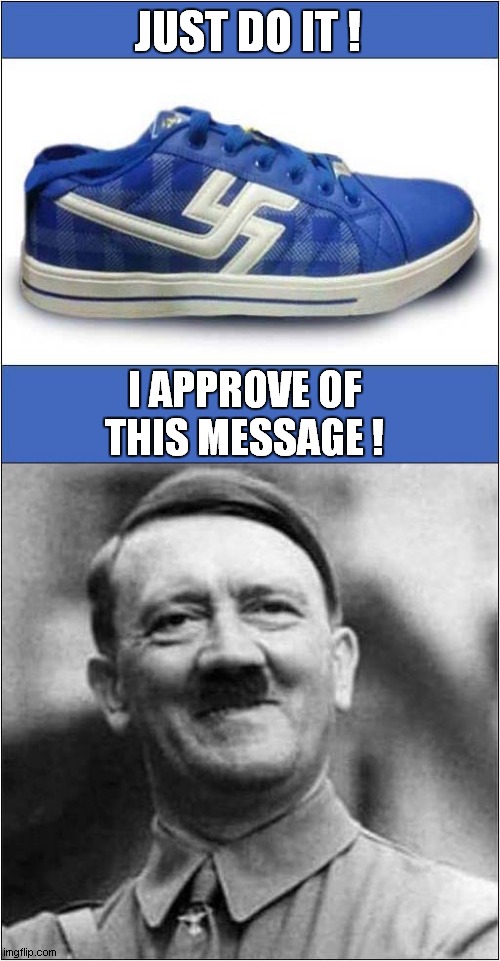} & \includegraphics[width=0.3\columnwidth,height=2.5cm]{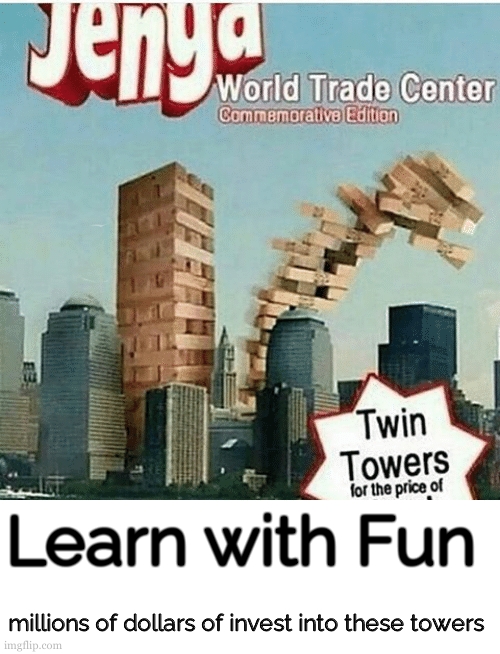} \\ \hline
GT & trainers, hitler & \begin{tabular}[c]{@{}c@{}}911 9/11 twin towers impact, \\ world trade center, jenga, \\ 9/11 truth movement\end{tabular} \\ \hline

\rowcolor[HTML]{CBCEFB} 
ST & hitler, trainers, nazi & 911, jenga, twin towers \\ \hline

R++ & blue, man, shoe , smile & building, city \\ \hline
R++(O) & \begin{tabular}[c]{@{}c@{}}Athletic shoe, Close-up,\\  Military person, Outdoor shoe\end{tabular} & Close-up, Toy block \\ \hline

R & blue , man , shoe & building, city \\ \hline

R(P) & -- &  --\\ \hline

R(O) & \begin{tabular}[c]{@{}c@{}}Athletic shoe, Black-and-white,\\   Military person, Outdoor shoe\end{tabular} & Toy block \\ \hline

T2T & \begin{tabular}[c]{@{}c@{}}shoe, picture, uniform, \\ person, man\textbackslash{}nUser Specified\end{tabular} & \begin{tabular}[c]{@{}c@{}}building, city,\\  game\textbackslash{}nUser Specified\end{tabular} \\ \hline

T2T(P) & \begin{tabular}[c]{@{}c@{}}
death, hitler, adolf hitler, \\
holocaust, family,\\ shoes\textbackslash nUser Specified
\end{tabular}
&  \begin{tabular}[c]{@{}c@{}}9/11, twin towers,\\ 911 9/11 twin towers \\impact\textbackslash nUser Specified\end{tabular} \\ \hline

\end{tabular}
\caption{\footnotesize Examples of tags generated by different models. 
GT:\textsc{Ground Truth}, ST: \ours{}, R: \textsc{RAM}, R++: \textsc{RAM++}, 
R(O): \textsc{RAM} open-set, R++(O): \textsc{RAM++} open-set, 
T2T: \textsc{Tag2Text}, R(P): \textsc{RAM} pre-trained, 
T2T(P): \textsc{Tag2Text} pre-trained using \datas{}. 
Please find more examples in Table~\ref{tab:model_compare_gen_tags_extra_examples} in the Appendix.}
\label{fig:model_compare_gen_tags}
\end{table}

\if{0}\section{Baselines:}
\label{main:baselines}

\subsection{Tag generation module}
\label{main:tag_generation_module_baselines}
\textsc{Tag2Text}, is a vision-language pre-training framework that incorporates image tagging to enhance the learning of visual-linguistic representations.  \textsc{Tag2Text} extracts tags directly from associated text, enabling the model to learn an image tagger while simultaneously guiding the vision-language learning process. The RAM model performs image tagging by training on large-scale image-text pairs. An initial model is trained by jointly learning from image captions and tags, supervised respectively by the original textual data and the parsed tags. This is followed by a data refinement engine that generates new tags and removes noisy annotations. The model on this cleaned dataset is retrained and further fine-tuned on a smaller, high-quality set. RAM++ is an open-set image tagging model that effectively harnesses multi-granular textual supervision. RAM++ unifies individual tag-level and global text-level supervision within a single alignment framework. To further enhance tag understanding, it leverages LLMs to transform semantically narrow tag supervision into broader, descriptive tag supervision, thereby enriching the model’s grasp of visual concepts in open-set scenarios. Both RAM and RAM++ models can either use their own trained tags or the open-set vocabulary of tags during inference, resulting in two different variants.\\
\noindent \\
\noindent\textbf{Metrics} : We use \textsc{chrf}, \textsc{meteor} and \metSS{} metrics to evaluate the similarity between the ground truth and the generated tags. 
\\

\noindent\textbf{Generation performance}: The results of tag generation are presented in Table~\ref{tab:tag_metrics}. The top rows by far outperforms the state-of-the-art baselines in terms all the evaluation metrics. Among our models, \ours{} has the best tag generation performance. 

\subsection{Toxic Meme Detection}
\label{main:baselines_toxic_meme_detection}
\fi
\if{0}
\section{Ablation}
\label{main:ablation_study}

In Table~\ref{tab:master_table_comparison}, we study the impact of various model improvements and make several key observations regarding the efficacy of our \ours{}, compared to other approaches.\\
\noindent The integration of generative tagging during training is found to promote the classification ability of the multi-tasking framework. These results indicate that the joint optimization of $\mathcal{L}_{cls}$ and $\mathcal{L}_{gen}$  as mentioned in \eqref{eq:total_loss} is significantly enhanced when the classification head is provided with the most confident internal representations. Our best approch \ours{} as defined in ~\eqref{main:entropy_guided_framework} successfully bypass noisy token embeddings. This demonstrates that the internal architecture may contain more valuable token embeddings than the final layer, which is traditionally followed. This approach yielding consistent improvements for both \sysP{} and \sysLV{} backbones(see the \textsc{LP} row of Table~\ref{tab:master_table_comparison}) with out affecting the tag generation performance much.\\
\noindent Additionally, we also checked the few-shot setting using the model \sysGPTO{}. Basically, we used CLIP for the exemplar selection process, and we evaluated the effectiveness of few-shot learning using \textbf{\textit{Tag embeddings}}: To obtain tag-level similarity, we first compute the CLIP embeddings for each individual tag associated with a sample. Next, we find the maximum similarity of each individual query tag embedding with the tag embeddings of an example meme. For instance, given a test sample with tags $\{t_1, t_2\}$ and an example meme with tags $\{e_1, e_2, e_3\}$, we compute the cosine similarity between the CLIP embedding of tag $t_1$ and those of $e_1$, $e_2$, and $e_3$, and record the maximum similarity value. We repeat the same process for $t_2$ and $t_3$. The final tag similarity between the test and the example sample is then defined as the mean of the maximum similarity scores. The examples with the highest mean scores are chosen as the few-shot examples. Overall, \ours{} multitasking approach yields better performance.\fi

\section{Error analysis}
\label{main:error_analysis}

\begin{table}[t]
\centering
\scriptsize
\setlength{\tabcolsep}{6pt}
\renewcommand{\arraystretch}{1.15}
\begin{tabular}{lcc}
\toprule
\textbf{category} & \textbf{\sysP{}} & \textbf{\sysLV{}} \\
\midrule
hateful    & 0.2288 $\pm$ 0.2088 & 0.3029 $\pm$ 0.1968 \\
dangerous  & 0.2423 $\pm$ 0.2043 & 0.3132 $\pm$ 0.1790 \\
normal     & 0.3221 $\pm$ 0.1670 & 0.3478 $\pm$ 0.1557 \\
offensive  & \textbf{0.4950 $\pm$ 0.1210} & \textbf{0.5559 $\pm$ 0.1065} \\
\bottomrule
\end{tabular}
\caption {Average uncertainty scores ($\pm$ standard deviation) across different toxicity categories for \sysP{} and \sysLV{}. Higher numbers signify higher uncertainty in the model's classification.}
\label{tab:avg_uncertainty_error_analysis}
\end{table}

To understand the failures of the model we conduct an error analysis that helps to identify how these models struggle in classification; due to systematic weaknesses in contextual reasoning. We perform two complementary analyses to characterize these errors. \\
\noindent \textbf{(i)} We identify the most frequent tags for which each model struggles to predict the correct classification labels. \noindent\textbf{Key observations:}
\noindent For \sysLV{}, the majority of the misclassifications are concentrated around tags corresponding to \textit{death-related events} (\texttt{death, funeral, suicide, murder, heart attack}), \textit{mental health and violence} (\texttt{depression, gun control, slavery}), and \textit{religious or cultural entities} (\texttt{religion, Jesus, Bible}), indicating a bias toward emotionally sensitive scenes.
\noindent In the case of \sysP{}, higher error frequencies are observed for tags associated with \textit{humor and internet slang} (e.g., \texttt{comic, lol, oof size large, cursed}), suggesting that the model is unable to effectively interpret implicit sarcasm and meme culture.\\

\noindent \textbf{(ii)} To better understand where the models fail, we group all false positive samples according to their incorrectly predicted classes and perform an uncertainty-based error analysis. We quantify the model's behavior using an uncertainty score derived from the Maximum Softmax Probability (MSP)~\cite{hendrycks2017baseline}. This uncertainty score is defined as:

\begin{equation}
\mathcal{U}(x) = 1 - \max_{c} p_{\theta}(c \mid x)
\end{equation}

\noindent
where \(x\) denotes a multimodal input sample, \(c \in \{\textit{hateful}, \textit{offensive}, \textit{dangerous}, \textit{normal}\}\) represents a toxicity class, and \(p_{\theta}(c \mid x)\) is the model-predicted posterior probability given input \(x\).

\noindent\textbf{Key observations:}
Table~\ref{tab:avg_uncertainty_error_analysis} shows that \sysLV{} consistently exhibits higher uncertainty for false positive predictions. Conversely, \sysP{} shows lower confidence margins. Both models display the highest uncertainty in the \emph{offensive} category, indicating difficulty in learning accurate patterns, whereas interestingly the lowest uncertainty is observed in the \emph{hateful} category.

\section{Conclusion}
\label{main:conclusion}
This work makes several key contributions toward advancing research in toxic meme detection and multimodal content moderation. \textbf{First}, we curate a richly annotated dataset of 6,300 real-world meme-based posts through a two-stage labelling process which we call \datas{}. \textbf{Second}, we enhance contextual understanding by introducing auxiliary metadata including meme titles and, most importantly, \cam{collaborative tags.} \textbf{Third}, we propose \ours{} that detects toxicity as well as generates tags given an arbitrary input meme. \textbf{Finally}, we compare the performance of toxicity detection and tag generation on an array of benchmarks and baselines for generalizability. Collectively, our contributions address key bottlenecks in semantically aligned toxic meme moderation and thus provides a foundation for building more accurate, context-aware, and socially responsible systems.

\section{Limitations}
\label{main:limitations}
While this work offers several important contributions, it also has a few limitations. \textit{\textbf{First}}, our dataset is limited to image-based memes, excluding other emerging modalities such as video and audio, which are increasingly used to disseminate harmful content. Future work will aim to extend this framework to multimodal datasets that better reflect current online communication trends.
\textit{\textbf{Second}}, we do not examine the psychological and emotional impact of hateful memes on viewers. Exposure to such content may contribute to anxiety, depression, or desensitization -- an important area that lies beyond the current scope. Addressing this limitation would require collaboration with psychologists, mental health experts, and affected communities.
\textit{\textbf{Third}}, we do not investigate the real-world consequences of online hate memes, particularly their potential to incite offline violence or criminal behaviour. Understanding this link between online toxicity and offline harm is a critical direction for future research.

\section*{Acknowledgments}
\label{main:acknowledgments}

This work was supported in part by the \textbf{Scheme for Promotion of Academic and Research Collaboration (SPARC)}, Ministry of Education, Government of India, and by the \textbf{Anusandhan National Research Foundation (ANRF)} through a Core Research Grant (CRG). We gratefully acknowledge their support.


\bibliography{main}
\bibliographystyle{acl_natbib}

\appendix

\section{Ethics statement}
\label{main:ethics_statement}
We strictly adhered to the policies of the social media platforms from which the dataset was curated. All memes and associated metadata were manually collected from publicly available content after agreeing to relevant copyright terms, and user anonymity was preserved throughout the process. During annotation, we followed detailed guidelines to safeguard annotators' mental well-being and conducted two-stage annotation with pilot studies to reduce subjective bias. We acknowledge the potential for bias in both annotation and model predictions; to mitigate this, we employed diverse evaluation metrics and experimental setups to assess model robustness. This study complies with ethical standards for research involving publicly available data, with a focus on transparency, privacy, and minimising harm. Our work aims to support the broader effort to combat online hate while remaining mindful of its own limitations.


\section{Prompt templates}
\label{main:prompt}
\ssubha{
We display below the prompts used for experiments.
}

\begin{tcolorbox}[colback=blue!5!white,colframe=blue!75!black,title=Prompts used for fine-tuning.]
\scriptsize
Generate tags by considering image, ocr and title\\
\end{tcolorbox}

\begin{tcolorbox}[colback=blue!5!white,colframe=blue!75!black,title=System prompt for zero-shot and few-shot classification tasks with VLMs.]
\scriptsize
  You are an AI assistant tasked with classifying memes.\\
Consider the following definitions.\\
$\bullet$~hateful: \{definition of hatful\}\\
$\bullet$~dangerous: \{definition of dangerous\}\\
$\bullet$~offensive: \{definition of offensive\}\\
$\bullet$~normal: \{definition of normal\}\\
Based on the above definitions, the input image, title and the extracted OCR
text from the image delimited by three backticks classify the meme as hateful, dangerous, offensive or normal.\\

Example output for hateful meme: \{hateful.\}\\
Example output for dangerous meme: \{dangerous.\}\\
Example output for offensive meme: \{offensive.\}\\
Example output for normal meme: \{normal\}
\end{tcolorbox}

\begin{tcolorbox}[colback=blue!5!white,colframe=blue!75!black,title=System prompt for zero-shot classification tasks with LLMS.]
\scriptsize
  You are an AI assistant tasked with classifying memes.\\
Consider the following definitions.\\
$\bullet$~hateful: \{definition of hatful\}\\
$\bullet$~dangerous: \{definition of dangerous\}\\
$\bullet$~offensive: \{definition of offensive\}\\
$\bullet$~normal: \{definition of normal\}\\
Based on the above definitions and  associated tags delimited by three backticks, classify the meme as hateful, dangerous, offensive or normal.\\

Example output for hateful meme: \{hateful.\}\\
Example output for dangerous meme: \{dangerous.\}\\
Example output for offensive meme: \{offensive.\}\\
Example output for normal meme: \{normal\}
\end{tcolorbox}


\section{Platforms}
\label{app:platforms}
\subsection{Data curation platform}
\label{app:curation_platform}

As stated earlier, we use \url{https://imgflip.com} as the curation platform for our data since it is centred on conversations using memes. We specifically use the \textit{streams}\footnote{\url{https://imgflip.com/streams}} feature of the platform to collect these memes. The selected streams and their descriptions are outlined as follows:

\noindent\textsc{\textbf{Dark\_Humour}} ($\sim$11k followers)\footnote{\url{https://imgflip.com/m/Dark_humour}} -- 
\textit{Welcome to Dark\_humour, Imgflip's premier community for offensive humour. Stream mood: Relax liberals, it's called dark humour.}

\noindent\textsc{\textbf{Memes\_Overload}} ($\sim$9.5k followers)\footnote{\url{https://imgflip.com/m/MEMES_OVERLOAD}} -- 
\textit{Hey there, welcome to MEMES\_OVERLOAD, Imgflip's 4th largest stream for memes! We're all here to have fun, so make sure to follow the rules, and keep on memeing on.}

\noindent\textsc{\textbf{Politics}} ($\sim$5k followers)\footnote{\url{https://imgflip.com/m/politics}} -- 
\textit{Humor and discussion around U.S. and world politics. Criticisms and debates are encouraged, but be constructive and don't harass anyone.}

\begin{figure}[!ht]
    \centering
    \includegraphics[width=1\linewidth]{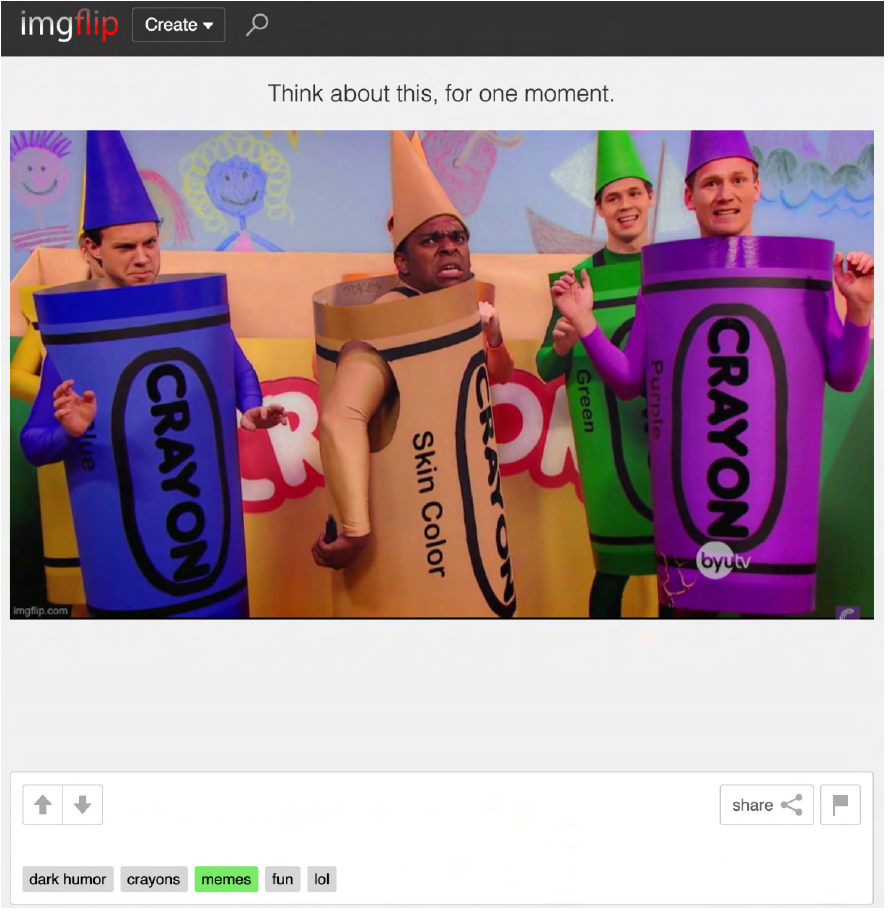}
    \caption{Post containing meme, title and tags from imgflip platform.}
    \label{fig:imgflip_platfrom}
\end{figure}
\noindent An example cropped post containing the corresponding title and tags is presented in Figure~\ref{fig:imgflip_platfrom}.


\section{Details of annotation}
\label{app:annotation_guidelines}
\begin{figure}[ht]
    \centering

    \begin{subfigure}{0.5\textwidth}
        \centering
        \includegraphics[width=\textwidth]{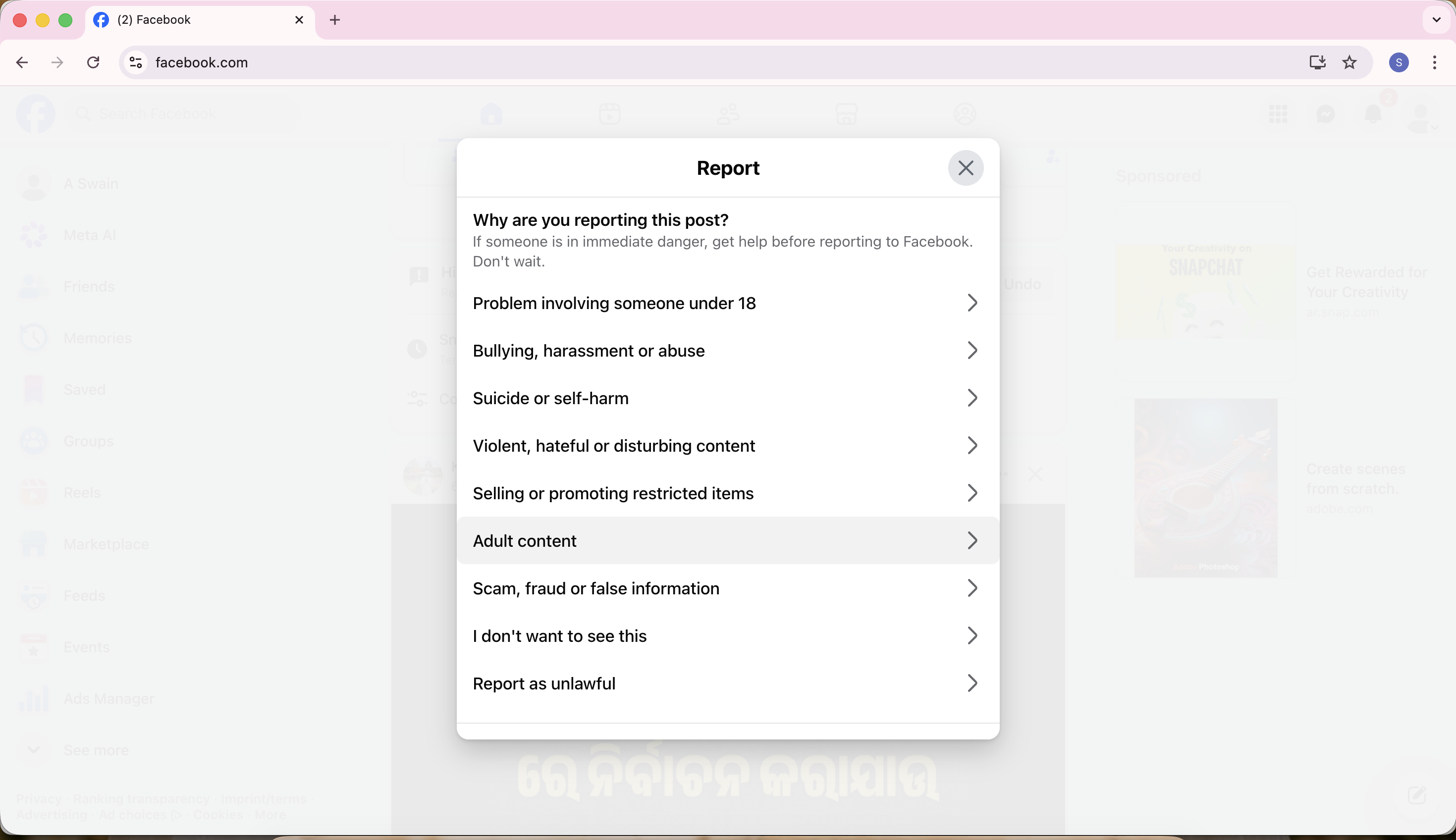}
        \caption{Facebook}
        \label{fig:facebook_policy}
    \end{subfigure}
    \hfill
    \begin{subfigure}{0.5\textwidth}
        \centering
        \includegraphics[width=\textwidth]{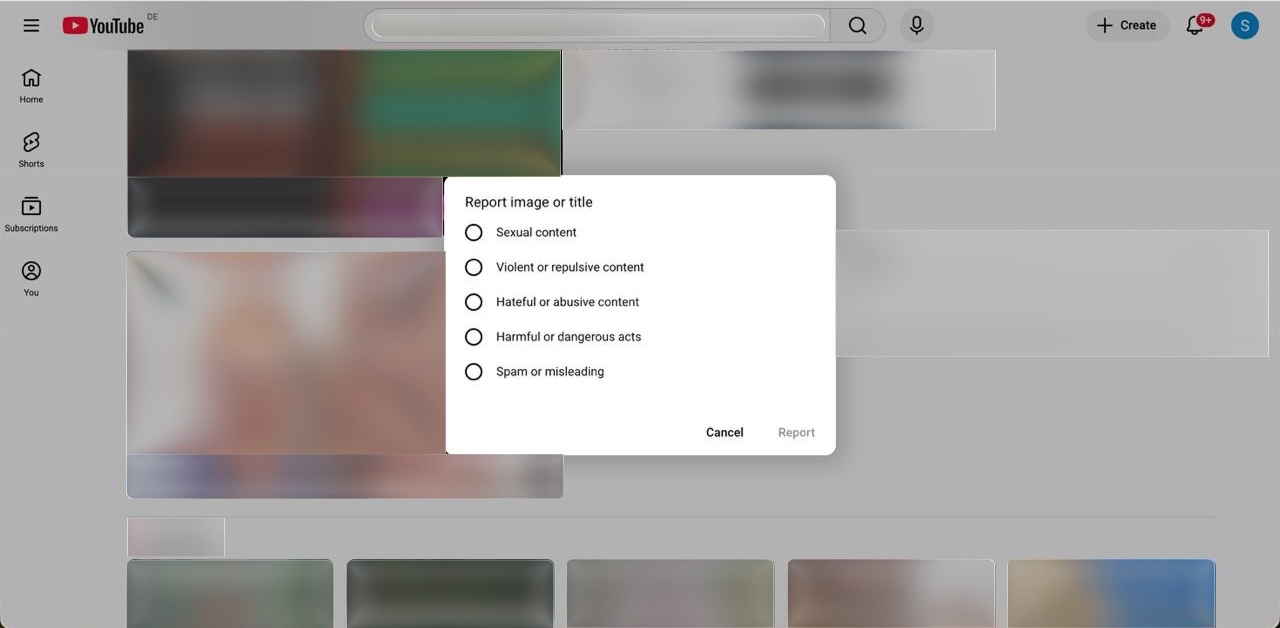}
        \caption{YouTube}
        \label{fig:youtube_policy}
    \end{subfigure}
    \hfill
    \begin{subfigure}{0.5\textwidth}
        \centering
        \includegraphics[width=\textwidth]{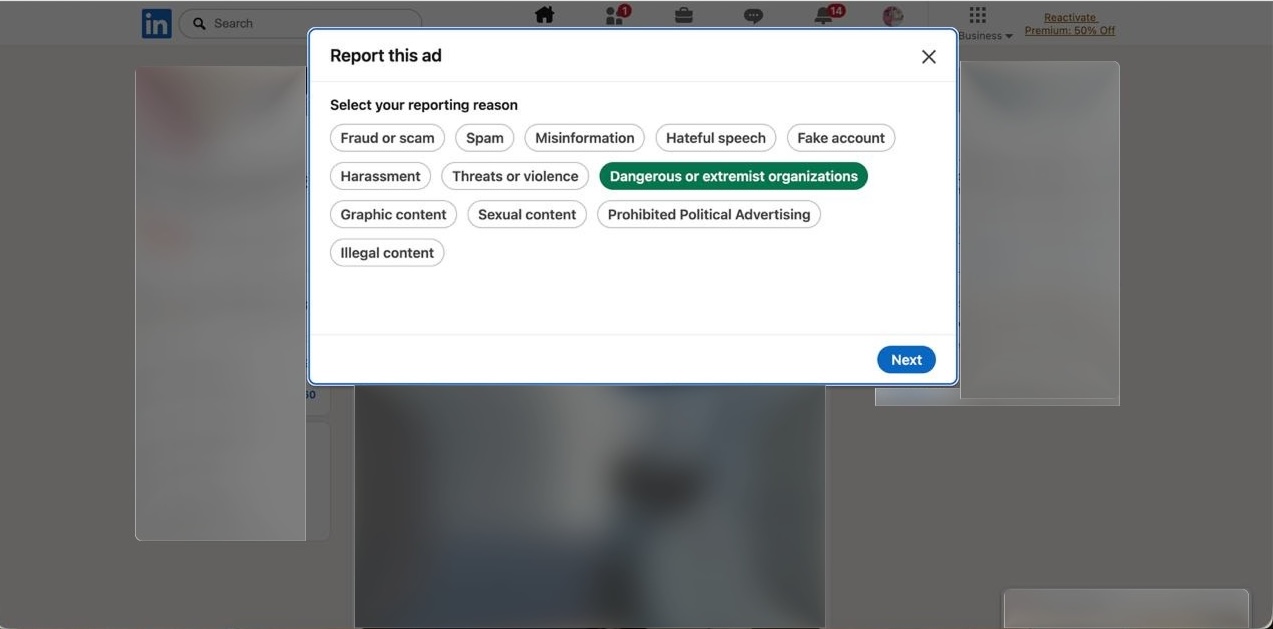}
        \caption{LinkedIn}
        \label{fig:linkedin_policy}
    \end{subfigure}

    \caption{Snapshots of reporting or policy interfaces from major social media platforms.}
    \label{fig:social_media_policies}
\end{figure}
Figure~\ref{fig:annotation_flowchart} presents a brief pipeline of the employed annotation process over two stages. The instructions that we designed are given below, and some samples from those provided to annotators are present in Table~\ref{tab:annotation_samples}. During the whole annotation process, annotators were asked to adhere to the definitions provided in the subsection~\ref{subsubsec:definitons}.

\noindent\textsc{\textbf{A}}-- Each annotator was provided with a separate PDF containing the meme, title, tags and the OCR extracted text from the meme. Along with that, a separate Google sheet was also provided to reduce annotation bias and ensure fairness.

\noindent\textsc{\textbf{B}}-- In Stage I, they were asked to strictly adhere to the provided definition of toxicity and segregate the memes as \textit{toxic} or \textit{normal}. Wherever confusion arose, annotators discussed in our daily scrum and through our Slack workspace.

\noindent\textsc{\textbf{C}}-- In Stage II, we provided the annotators with \textit{toxic} memes based on majority voting as per Stage I. Note that before starting Stage II they were fairly aware of the type of content moderation required on these social media platforms. As a first step, they were asked to segregate hateful content, then from the remaining to identify dangerous samples; finally left with offensive samples, which were also verified. All the annotations were performed by strictly adhering to the definition of \textit{hateful}, \textit{dangerous} and \textit{offensive} labels.

\begin{table*}[!t]
\centering
\scriptsize
\setlength{\tabcolsep}{2pt}
\renewcommand{\arraystretch}{1.2}

\begin{tabular}{p{2cm}|c|c}
\hline
\textbf{Image} &
\includegraphics[width=0.22\textwidth,height=2.4cm,keepaspectratio]{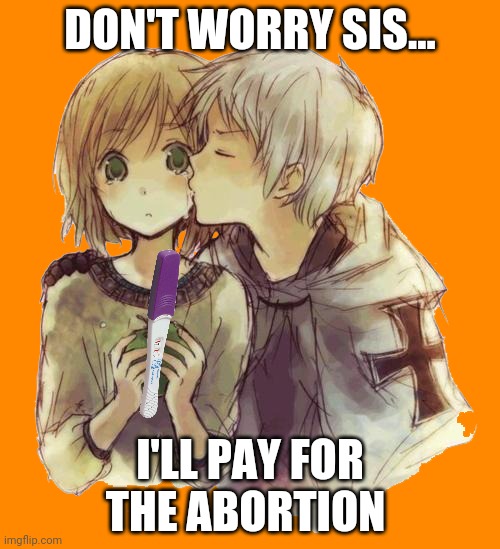} &
\includegraphics[width=0.22\textwidth,height=2.4cm,keepaspectratio]{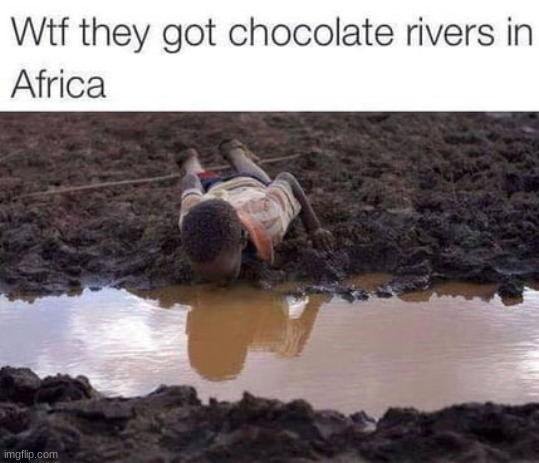} \\
\hline

ground truth &
alabama, incest, baby &
chocolate, rivers, africa \\
\hline

\rowcolor[HTML]{CBCEFB}
\ours{} &
alabama, problems, abortion, pregnancy &
chocolate, rivers, africa \\
\hline

RAM++ &
anime, brush, girl, person, kiss, love &
\begin{tabular}[c]{@{}c@{}}
boy, trench, floor, person, land,\\
lay, man, mud, puddle, water
\end{tabular} \\
\hline

RAM++ (O) &
Close-up &
Jeans \\
\hline

RAM &
anime, girl, person, kiss &
\begin{tabular}[c]{@{}c@{}}
boy, floor, person, land, lay,\\
man, mud, puddle, water
\end{tabular} \\
\hline

{RAM (P)} & --
 &--
 \\
\hline

RAM (O) &
-- &
Ribs \\
\hline

\textsc{Tag2Text} &
\begin{tabular}[c]{@{}c@{}}
anime, girl, people, person\\
User Specified
\end{tabular} &
\begin{tabular}[c]{@{}c@{}}
mud, puddle, water, man\\
User Specified
\end{tabular} \\
\hline

\textsc{Tag2Text (P)} &
women, psychopath\textbackslash nUser Specified&repost, christmas\textbackslash nUser Specified\\
\hline
\end{tabular}

\caption{Comparative examples of tags generated by different models. 
(O): open-set, (P): pre-trained using \datas{}.}
\label{tab:model_compare_gen_tags_extra_examples}
\end{table*}

\begin{table*}[!t]
\centering
\scriptsize
\setlength{\tabcolsep}{2pt}
\renewcommand{\arraystretch}{1.2}

\begin{tabular}{p{1.6cm}|c|c|c|c}
\hline
\textbf{Image} &
\includegraphics[width=0.20\textwidth]{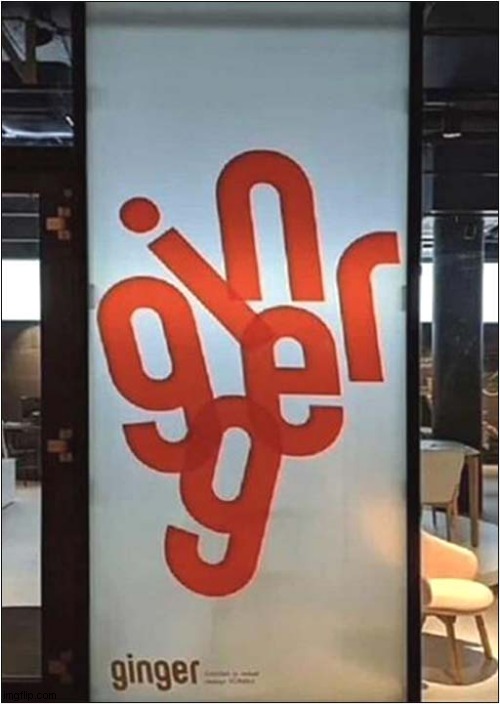} &
\includegraphics[width=0.20\textwidth]{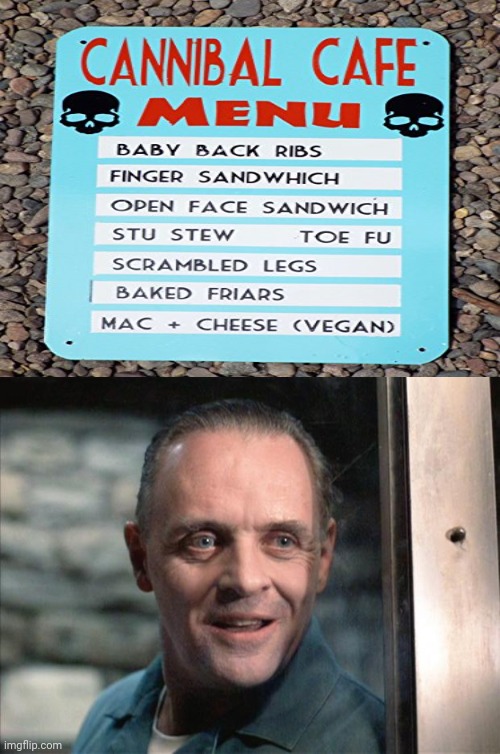} &
\includegraphics[width=0.20\textwidth]{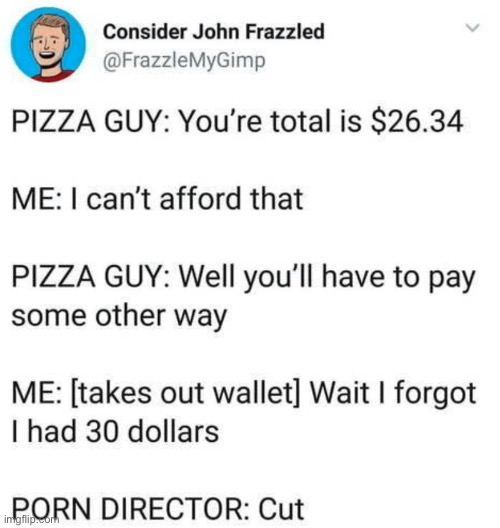} &
\includegraphics[width=0.20\textwidth]{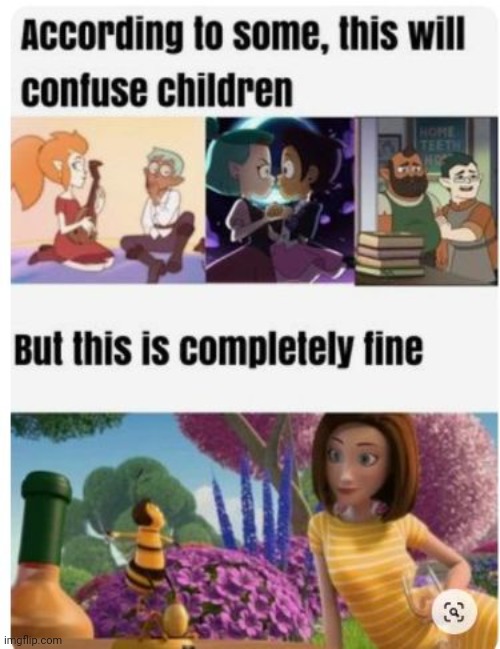} \\
\hline

\textbf{Title} &
\begin{tabular}[c]{@{}c@{}}
Does Anyone Else See The\\
Problem With This\\
Advertising Sign?
\end{tabular} &
Cannibal cafe menu &
If ykyk &
My friend sent me this \\
\hline

\textbf{Tags} &
signs, advertising &
\begin{tabular}[c]{@{}c@{}}
hannibal lecter, cannibals,\\
cannibal, cannibalism
\end{tabular} &
if you know you know, porn &
\begin{tabular}[c]{@{}c@{}}
if you know you know,\\
the owl house, bee movie
\end{tabular} \\
\hline

\textbf{OCR extracted text} &
imaflip.com ginger === S &
\begin{tabular}[c]{@{}c@{}}
CANNIBAL CAFE MENU BABY\\
BACK RIBS FINGER SANDWICH\\
OPEN FACE SANDWICH STU STEW\\
TOE FU SCRAMBLED LEGS\\
BAKED FRIARS MAC CHEESE (VEGAN)
\end{tabular} &
\begin{tabular}[c]{@{}c@{}}
Consider John Frazzled\\
FrazzleMyGimp PIZZA\\
GUY: Your total is \$26.34\\
ME: I can't afford that\\
PIZZA GUY: Pay another way\\
PORN DIRECTOR: Cut
\end{tabular} &
\begin{tabular}[c]{@{}c@{}}
According to some,\\
this will confuse children\\
But this is completely fine\\
HOME TEETH 3
\end{tabular} \\
\hline

\textbf{Expert annotation} &
\textbf{hateful} &
\textbf{dangerous} &
\textbf{offensive} &
\textbf{normal} \\
\hline
\end{tabular}

\caption{Four memes from the expert annotation samples provided to the annotators.}
\label{tab:annotation_samples}
\end{table*}

\begin{table*}[!t]
\centering
\setlength{\tabcolsep}{6pt}
\renewcommand{\arraystretch}{1.1}

\begin{tabular}{cc}
\includegraphics[width=0.45\textwidth,keepaspectratio]{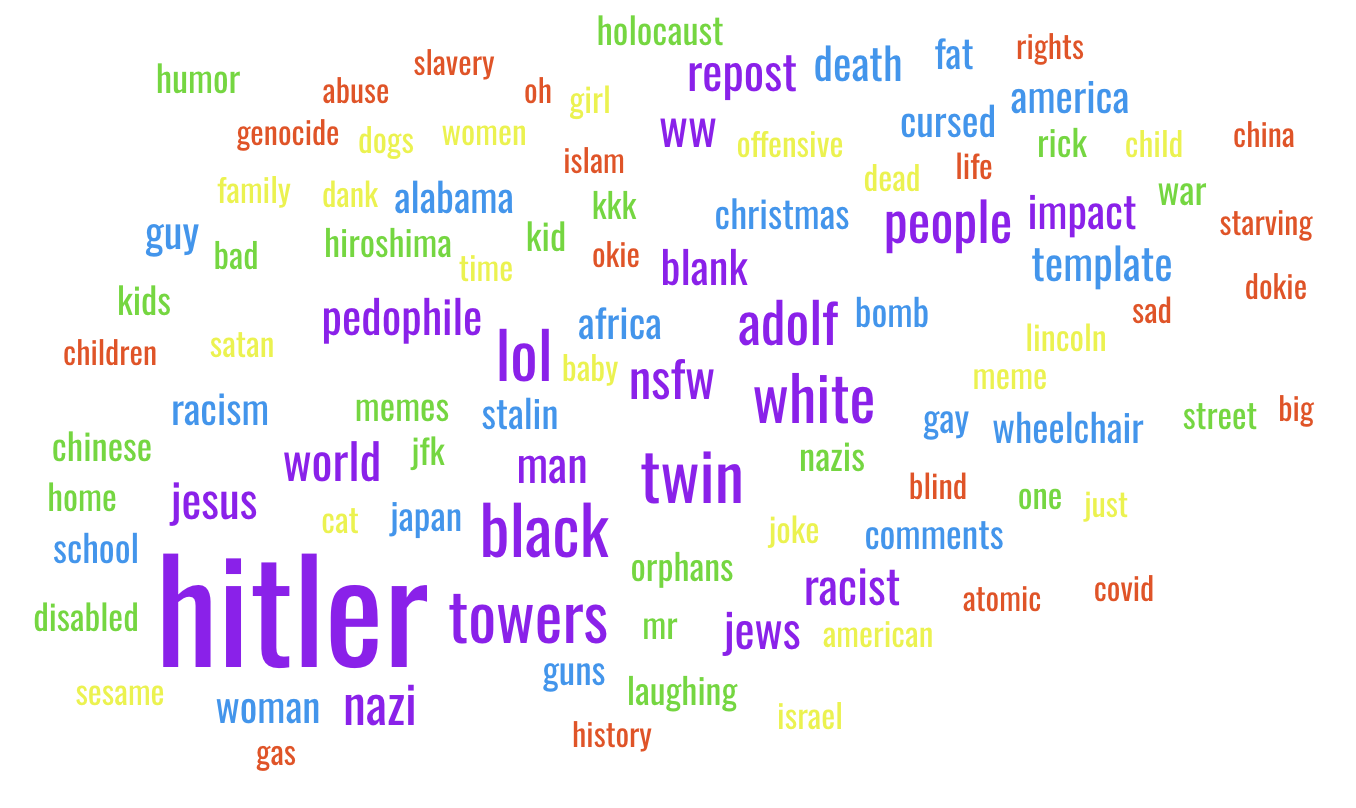} &
\includegraphics[width=0.45\textwidth,keepaspectratio]{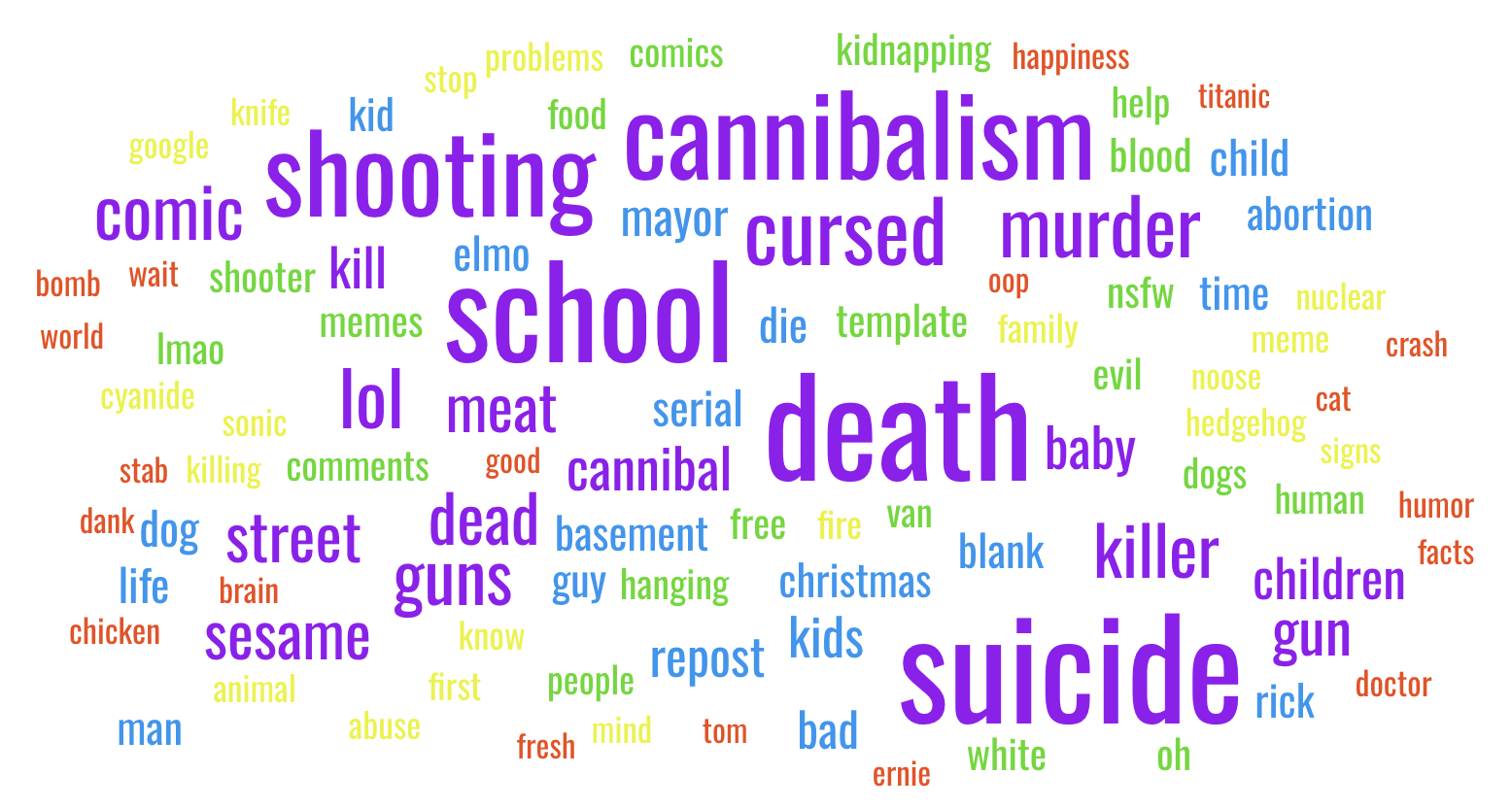} \\
\textbf{hateful} & \textbf{dangerous} \\[0.6em]

\includegraphics[width=0.45\textwidth,keepaspectratio]{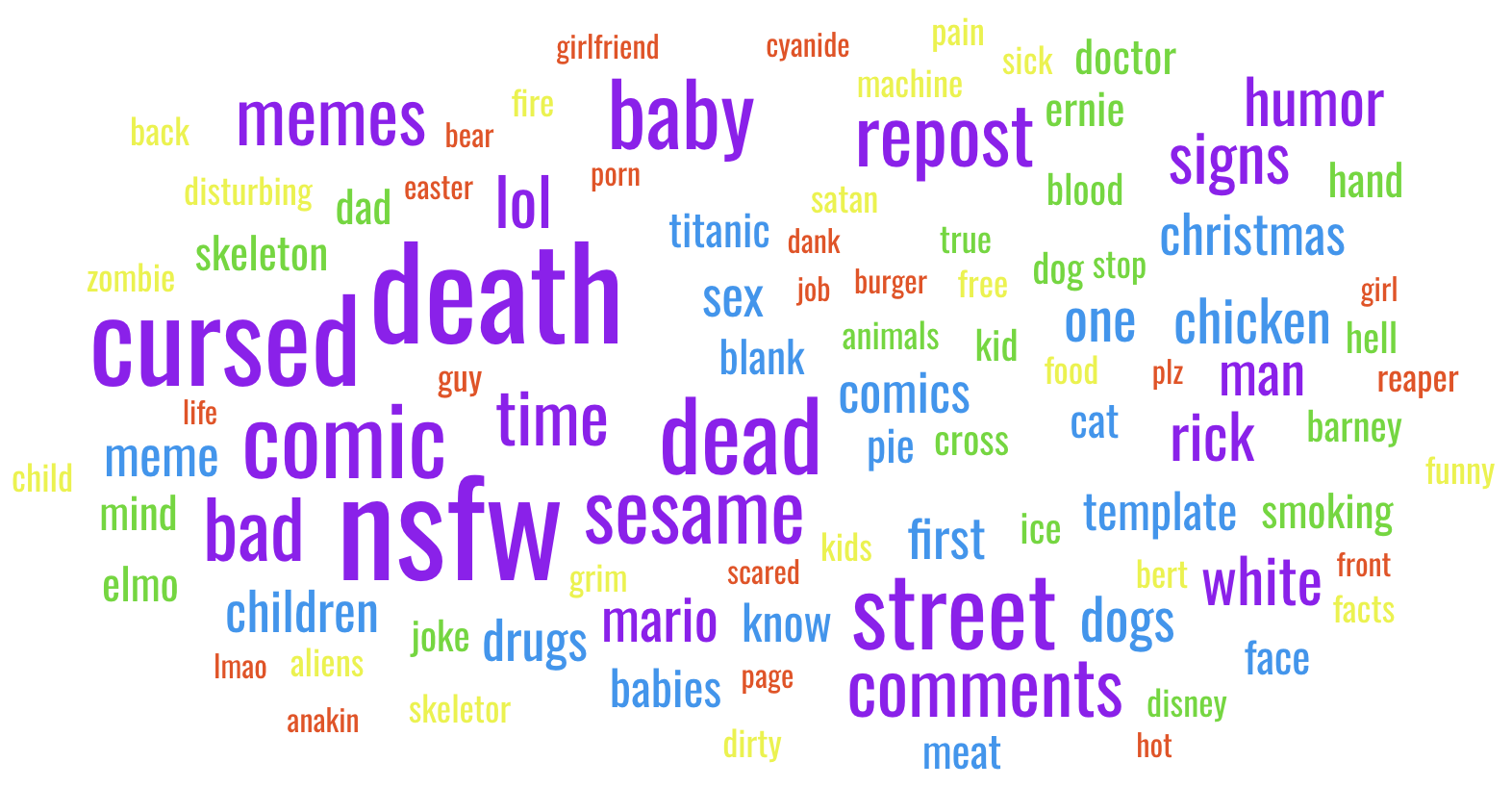} &
\includegraphics[width=0.45\textwidth,keepaspectratio]{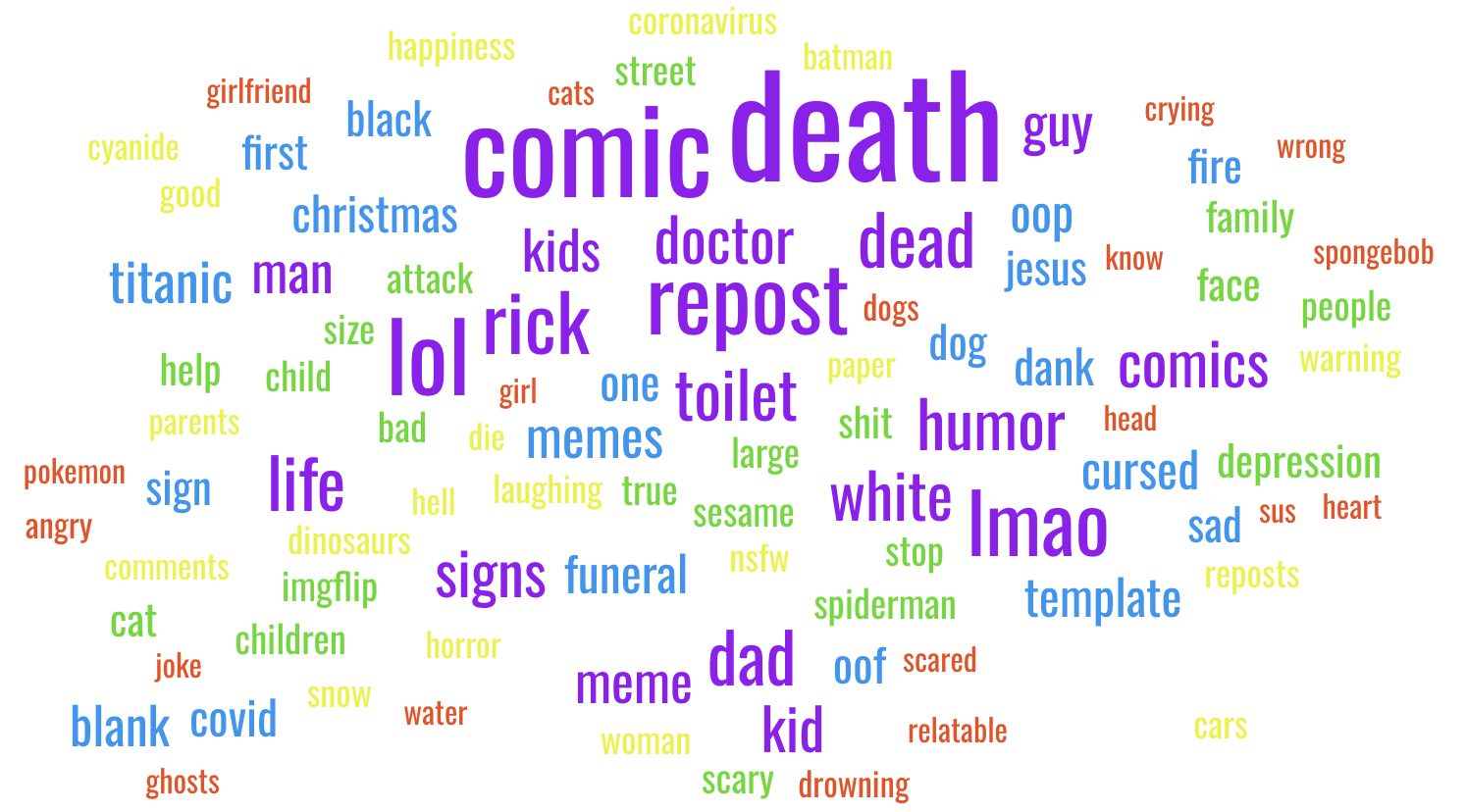} \\
\textbf{offensive} & \textbf{normal}
\end{tabular}

\caption{Label-wise word cloud of tags.}
\label{fig:word_clouds}
\end{table*}

\subsection{Definitions}
\label{subsubsec:definitons}
\textbf{hateful} -- \textit{Reference}:~\cite{kiela2020hateful} -- A direct or indirect attack on people based on characteristics, including ethnicity, race, nationality, immigration status, religion, caste, sex, gender identity, sexual orientation, and disability or disease. 'Attack' is defined as violent or dehumanising (comparing people to non-human things, e.g., animals) speech, statements of inferiority, and calls for exclusion or segregation. Mocking hate crime is also considered hateful.\\
\textbf{dangerous} -- \textit{Reference\footnote{\url{https://www.dangerousspeech.org/dangerous-speech}}} -- A text, meme or speech which is not hateful but uses any form of expression that can increase the risk of its audience to condone or participate in violence against members of another group will be considered dangerous.\\
\textbf{offensive} -- \textit{References}:~\cite{roy2023probing, Mathew_Saha_Yimam_Biemann_Goyal_Mukherjee_2021} -- A text, meme or speech which is neither hateful nor dangerous but uses abusive slurs or derogatory terms will be considered offensive.\\
\textbf{toxic} -- \textit{Reference}: PerspectiveAPI\footnote{\href{https://developers.perspectiveapi.com/s/about-the-api-model-cards?language=en_US}{Perspective API Model Card}} -- A rude, disrespectful, or unreasonable comment that is likely to make you leave a discussion.\\
\textbf{normal} -- A meme which is not toxic and follows social norms.

\subsection{Safety guidelines}
Primarily, we performed the following steps to keep our annotators mentally safe with such content:\\
\textbf{(i)} Only 50 samples per day were provided to them, wherein we updated the PDF and added corresponding samples in the Google sheet.\\
\textbf{(ii)} We conducted 15 minutes of daily mental well-being sessions in our daily scrum by adopting various online activities suggested by WHO. \footnote{https://www.who.int/news-room/feature-stories/mental-well-being-resources-for-the-public}\\
\textbf{(iii)} We also asked the annotators to agree to the data source platform's terms and policies. Further, we strictly ensured that they did not disclose the identity of users in the provided Google sheet.


\section{Further analysis of the dataset}\label{sec:more_analysis}
\begin{table*}[!ht]
\centering
\scriptsize
\setlength{\tabcolsep}{4pt}
\renewcommand{\arraystretch}{1.2}

\begin{tabular}{c|c|c|c}
\hline
\textbf{Image} &
\includegraphics[width=0.28\textwidth]{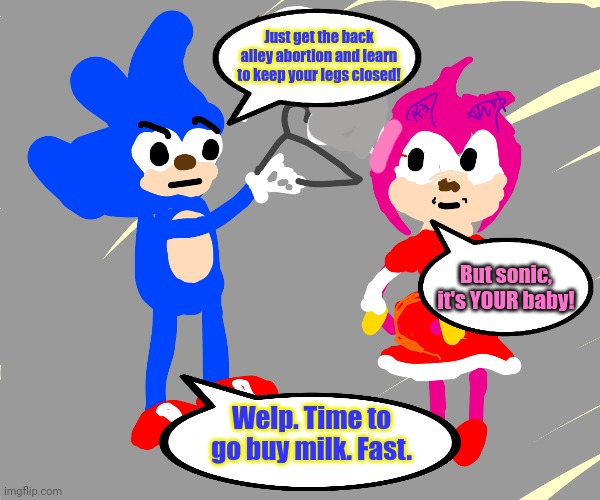} &
\includegraphics[width=0.28\textwidth]{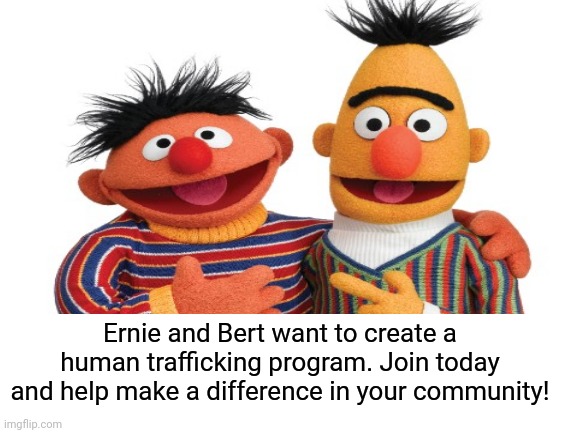} &
\includegraphics[width=0.28\textwidth]{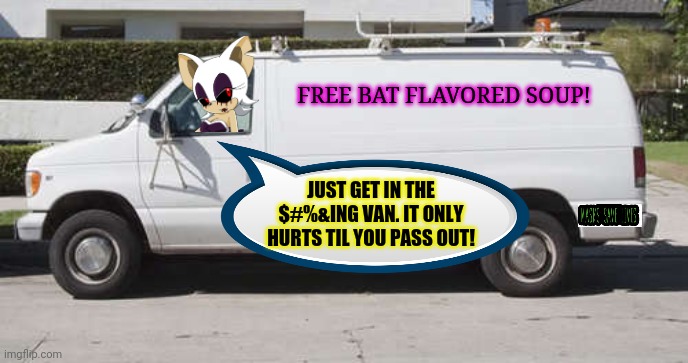} \\
\hline

\textbf{Title}
Gotta go fast &
All members get 69\% off all items &
Rouge.exe is getting desperate! \\
\hline

\textbf{Tags} &
\begin{tabular}[c]{@{}c@{}}
gotta go fast, sonic\\
the hedgehog, \textbf{abortion}
\end{tabular} &
\begin{tabular}[c]{@{}c@{}}
\textbf{human stupidity}, ernie and bert,\\
\textbf{traffic light}
\end{tabular} &
\begin{tabular}[c]{@{}c@{}}
rougeexe, sonic the hedgehog,\\
\textbf{free candy van}
\end{tabular} \\
\hline

\textbf{\begin{tabular}[c]{@{}c@{}}OCR extracted\\ text\end{tabular}} &
\begin{tabular}[c]{@{}c@{}}
Just get the back alley abortion and\\
learn to keep your legs closed! Welp.\\
Time to go buy milk. Fast. But\\
sonic, it's YOUR baby!
\end{tabular} &
\begin{tabular}[c]{@{}c@{}}
Ernie and Bert want to create\\
a human trafficking program.\\
Join today and help make a\\
difference in your community!
\end{tabular} &
\begin{tabular}[c]{@{}c@{}}
FREE BAT FLAVORED SOUP!\\
JUST GET IN THE \$\#\%\&ING VAN.\\
IT ONLY HURTS TIL YOU PASS OUT!\\
MASKS SAVE LIVES
\end{tabular} \\
\hline

\textbf{\begin{tabular}[c]{@{}c@{}}Ground truth\\ annotation\end{tabular}} &
\textbf{dangerous} &
\textbf{dangerous} &
\textbf{dangerous} \\
\hline
\end{tabular}

\caption{\footnotesize Dangerous memes: portrayal of danger through anime/cartoon characters.}
\label{tab:example_memes_analysis}
\end{table*}

\noindent\textbf{Word clusters of tags}: Figure~\ref{fig:word_clouds} presents the word cloud of tags for \textit{hateful}, \textit{dangerous}, \textit{offensive} and \textit{normal} memes. We can clearly segregate the word clouds for hateful and dangerous, hence signifying our initial observations. For offensive and normal, even though they have similar words, the presence of terms like \textit{cursed, nsfw} and usage of comic character terms like \textit{sesame, street} significantly separates them.

\section{Employed models}
\label{app:models}
\ssubha{
\noindent\textbf{\sysP{}}: We use \sysP{}-2 which presents a series of models of varying size with further improved capabilities due to the incorporation of \textsc{Gemma-2} LLM instead of \textsc{Gemma}. We use \texttt{google/paligemma2-10b-pt-224} version from HuggingFace in this work.\\
\noindent\textbf{\sysLV{}}: We use \sysLV{}-1.5~\cite{Liu_2024_CVPR} that has shown significant improvement over its prior models. We use \texttt{llava-hf/llava-1.5-7b-hf} checkpoint for running our experiments.\\
\noindent\textbf{\sysGPTO{}}: One of the first models by OpenAI team to have multimodal capability, \sysGPTO{}~\cite{openaiGPT4} has proved its wide applicability across multiple domains\footnote{\url{https://learn.microsoft.com/en-us/azure/ai-services/openai/}}.\\
\textbf{\textsc{Mistral}}: \textsc{Mistral} is an instruction-tuned large language model designed for efficient reasoning and strong performance across diverse NLP tasks. We use \texttt{ mistralai/Mistral-7B-Instruct-v0.3} checkpoint from Huggingface.\\
\textbf{\textsc{Llama}}: We use \textsc{Llama-3.1}is which is a part of Meta’s LLaMA series, designed for high-quality and robust generalization. We use \texttt{meta-llama/Llama-3.1-8B-Instruct}.
\noindent\textbf{License agreement}: We agreed to the terms of usage of all employed VLMs before using them in our work.
}

\section{Experimental setup}
\label{app:experimental_setup}

\noindent We train both the uni- and multi-tasking framework using a consistent experimental configuration to ensure a fair comparison. For the multi-task setting, the training objective consists of two components: a weighted cross-entropy loss for the classification task and the default autoregressive generation loss for toxic tag generation. In the case of uni-task learning, there is no generation loss. For computing class weights, we use the following weighting scheme: \href{https://www.geeksforgeeks.org/machine-learning/how-does-the-classweight-parameter-in-scikit-learn-work/}. All input sequences are truncated or padded to a fixed maximum sequence length of 2048 tokens, and the random seed is fixed to 42 to ensure reproducibility. \\
 Training is performed for three epochs with a per-device batch size of 1 and a gradient accumulation step of 4, resulting in an effective batch size of 4. We use the \texttt{AdamW} optimizer with a learning rate of $2 \times 10^{-5}$, weight decay of $1 \times 10^{-6}$, and $\beta_2$ set to 0.999, along with a warm-up of two steps. For memory-efficient training, the model is loaded using 8-bit quantization via the \texttt{BitsAndBytes} configuration, with computations carried out in \texttt{FP16} precision. For the multi-task Learning framework, we apply LoRA-based parameter-efficient fine-tuning with rank 8, targeting the following modules: \{\texttt{q\_proj}, \texttt{k\_proj}, \texttt{v\_proj}, \texttt{o\_proj}, \texttt{gate\_proj}, \texttt{up\_proj}, \texttt{down\_proj}\}.

In contrast, for the uni-task learning framework, the model backbone is fully frozen during training. Extracted hidden representations are normalized using the default normalization layers provided by the backbone model at the time of forward propagation. A sequence compression module is used to extract features into a fixed-dimensional embedding, which is then passed through a label classification head to predict the final class labels.\\
\noindent For the \datas{} dataset, we fix the number of classes to four while for the \fhm{} and \mami{} datasets, we fix the number of classes to two. For result calculation and analysis of stage I, we merge the hateful, dangerous, and offensive categories into a single toxic class. This aggregation is the reverse of the two-stage annotation strategy followed during dataset construction. \\

\end{document}